\documentclass[conference]{IEEEtran}

\IEEEoverridecommandlockouts
\def\BibTeX{{\rm B\kern-.05em{\sc i\kern-.025em b}\kern-.08em
    T\kern-.1667em\lower.7ex\hbox{E}\kern-.125emX}}
\usepackage{hyperref}
\usepackage{url}
\usepackage{booktabs}       % professional-quality tables
\usepackage{amsfonts}       % blackboard math symbols
\usepackage{nicefrac}       % compact symbols for 1/2, etc.
\usepackage{microtype}      % microtypography
\usepackage{multirow}
\usepackage{marvosym}
\usepackage{color}
\usepackage{threeparttable}
\usepackage[ruled,linesnumbered]{algorithm2e} 
\usepackage{siunitx} % Required for alignment
\usepackage{amsthm}
\usepackage{verbatim}
\usepackage{xcolor}
\usepackage{colortbl}
\usepackage{algorithmic}
\usepackage{bibentry}
\usepackage{subfigure}
\usepackage{multirow}
\usepackage{wrapfig}
\usepackage{graphicx}
\usepackage{epsfig}
\usepackage{color}
\usepackage{epstopdf}
\usepackage{rotating}
\usepackage{float}
\usepackage{multicol}
\usepackage{amssymb}
\usepackage{amsmath}
\usepackage{thmtools}
\usepackage{graphicx}
\usepackage{balance}
\usepackage{microtype}
\usepackage{graphicx}
\usepackage{subfigure}
\usepackage{multirow}
\usepackage{bm}
\usepackage{mathtools}
\usepackage{etoolbox}
\usepackage{cases}
\usepackage{enumitem}
\usepackage{xcolor}
\usepackage{marvosym}
\usepackage{subfigure}
\usepackage{color}
\usepackage{caption} 
\definecolor{backgreen}{RGB}{213, 232, 212}
\newcommand{\backg}{\cellcolor{backgreen!50}}
\usepackage{arydshln}

\newtheorem{theorem}{Theorem}

\newtheorem{proposition}{Proposition}

\newcommand{\boldres}[1]{{\textbf{\textcolor{red}{#1}}}}
\newcommand{\secondres}[1]{{\underline{\textcolor{blue}{#1}}}}
\definecolor{brown}{RGB}{139,64,0}
\renewcommand*{\thefootnote}{\fnsymbol{footnote}}
\usepackage[most]{tcolorbox}
\newtcbtheorem[auto counter, number within = section]{cmt}{Comment}{
	colbacktitle = black!60!white, colframe = black!60!white,
	colback = black!5!white,
	fonttitle=\bfseries,%fontupper=\itshape,
}{t}
\definecolor{reviseblue}{HTML}{00AEEF}

\begin{document}
\setcounter{equation}{0}
\setcounter{theorem}{0}
\setcounter{proposition}{0}
\setcounter{section}{0}
\setcounter{table}{0}
% \title{Urban-SSM: Domain-Adaptive Selective State Space Model for Efficient Spatio-Temporal Prediction}
\title{Damba-ST: Domain-Adaptive Mamba for Efficient Urban Spatio-Temporal Prediction} 

\author{
	\IEEEauthorblockN{
		Rui An\textsuperscript{1,2}, 
		Yifeng Zhang\textsuperscript{2}, 
		Ziran Liang\textsuperscript{2}, 
        Wenqi Fan\textsuperscript{2}\textsuperscript{\Letter}, 
        Yuxuan Liang\textsuperscript{3}, 
		Xuequn Shang\textsuperscript{1}\textsuperscript{\Letter}, 
        Qing Li\textsuperscript{2}} 
    \IEEEauthorblockA{\textit{\textsuperscript{1}Northwestern Polytechnical University}\\
    \textit{\textsuperscript{2}The Hong Kong Polytechnic Univeristy}\\
    \textit{\textsuperscript{3}The Hong Kong University of Science and Technology (Guangzhou)}}
    
	\IEEEauthorblockA{
    Email: \{rui77.an, yifeng.zhang, ziran.liang\}@connect.polyu.hk, \\wenqifan03@gmail.com, yuxliang@outlook.com, shang@nwpu.edu.cn, csqli@comp.polyu.edu.hk\\
    }
    \thanks{\textsuperscript{\Letter}Corresponding authors.
    }
}

\maketitle

\begin{abstract}
Training urban spatio-temporal foundation models that generalize well across diverse regions and cities (i.e., cross-domain scenarios) is critical for deploying urban services in unseen or data-scarce regions. Recent studies have typically focused on fusing cross-domain spatio-temporal data to train unified Transformer-based models. However, these models suffer from quadratic computational complexity and high memory overhead, limiting their scalability and practical deployment. Inspired by the efficiency of Mamba, a state-space model with linear time complexity, we explore its potential for efficient urban spatiotemporal prediction. However, directly applying Mamba as a spatio-temporal backbone leads to negative transfer and severe performance degradation in unseen regions. This is primarily due to inherent spatio-temporal heterogeneity and the recursive mechanism of Mamba's hidden state updates, which limit cross-domain generalization. To overcome these challenges, we propose Damba-ST, a novel domain-adaptive Mamba-based model for efficient urban spatio-temporal prediction. Damba-ST retains Mamba's linear complexity advantage while significantly enhancing its adaptability to heterogeneous domains. Specifically, we introduce two core innovations: (1) a Domain-Adaptive State Space Model that partitions the latent representation space into a shared subspace for learning cross-domain commonalities and independent, domain-specific subspaces for capturing intra-domain discriminative features; (2) three distinct Domain Adapters, which serve as domain-aware proxies to bridge disparate domain distributions and facilitate the alignment of cross-domain commonalities. Extensive experiments demonstrate the generalization and efficiency of Damba-ST. It achieves state-of-the-art performance on prediction tasks and demonstrates strong zero-shot generalization, enabling seamless deployment in new urban environments without extensive retraining or fine-tuning. 
\end{abstract}
% \footnotetext[1]{Corresponding authors.}

\maketitle

\section{Introduction}
\label{sec:Introduction}
Urban spatio-temporal systems describe the dynamic evolution of urban activities, including human mobility and vehicle dynamics (e.g., traffic flow, speed, taxi demand, and bike trajectories)~\cite{zheng2014urban,222}. These dynamics are characterized by both temporal and spatial dimensions, exhibiting variations in temporal volumes at different intersections~\cite{jin2023large,111,wu2019graph}. Accurate spatio-temporal traffic prediction is crucial for optimizing urban infrastructure, alleviating potential congestion and delays, and enabling prompt responses to emergencies~\cite{xie2020urban,cirstea2022towards,9101507}. Despite notable advancements in existing spatio-temporal modeling methods~\cite{zheng2020gman, luo2024lsttn,zhang2025robust,liang2025itfkan}, these models are typically designed for a specific domain, limiting their ability to generalize effectively to unfamiliar traffic environments.
\begin{figure*}[t]
\vskip -0.1in
\begin{center}
\hspace{-0.5cm}
\subfigure[Temporal Heterogeneity]{
\includegraphics[width=0.35\textwidth]{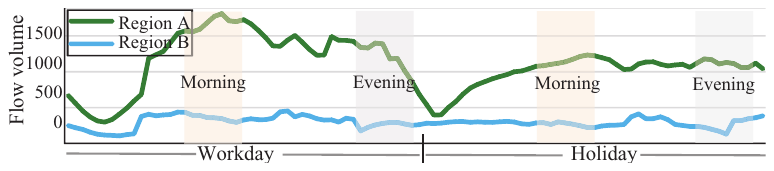}
\label{fig:Temporal Heterogeneity}
}
% \hspace*{\fill}
\hspace{-0.5cm}\subfigure[Spatial Heterogeneity]{
\includegraphics[width=0.19\textwidth]{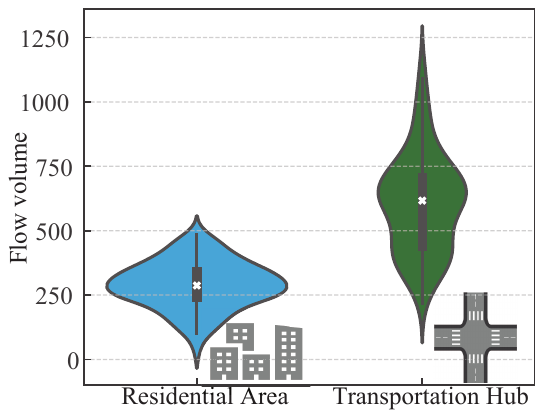}
\label{fig:Spatial Heterogeneity}
}
\hspace{-0.3cm}\subfigure[Diverse Data Distributions]{
\includegraphics[width=0.22\textwidth]{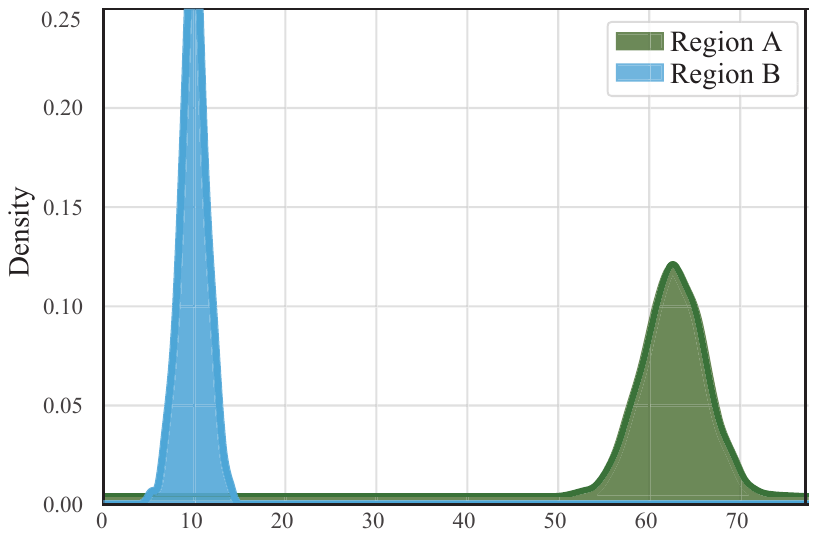}
\label{fig:Data Distributions}
}
% \vskip -0.08in
\subfigure[Partition Representation Space]{
\hspace{-0.2cm}\includegraphics[width=0.21\textwidth]{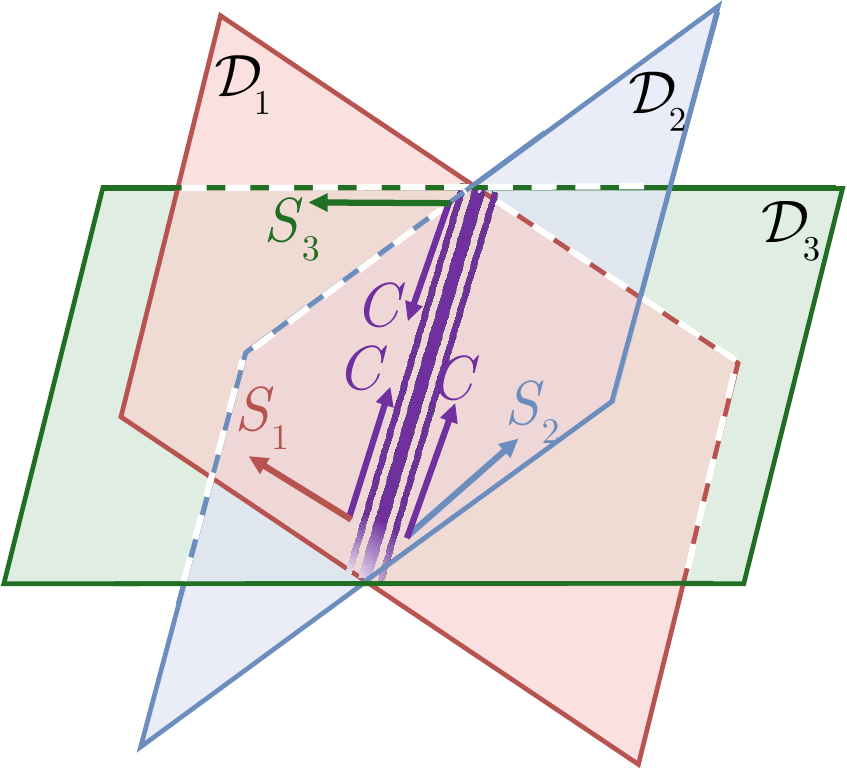}
\label{fig:subspace}
}
% \vskip -0.1in
\caption{\textbf{Spatio-Temporal Heterogeneity and Our Motivation}. (a) Regions A and B represent a transportation hub and residential area, each with distinct urban functions. Traffic patterns vary significantly over time (e.g., morning to evening and weekdays to holidays). (b) Distinct regions exhibit notable differences in traffic flow volume. (c) Such spatio-temporal heterogeneity leads to significant discrepancies in dataset distributions. (d) Given cross-domain data $\{\mathcal{D}_1,\mathcal{D}_2, \mathcal{D}_3\}$,  we propose partitioning the Mamba representation space into a shared subspace for cross-domain commonalities, denoted as $C$, and independent subspaces for domain-specific patterns, i.e., $\{S_1, S_2, S_3\}$.
}
\label{fig:motivation}
\end{center}
\vskip -0.2in
\end{figure*}

Over the past few years, pretrained foundation models have achieved significant success in Natural Language Processing (NLP) and Computer Vision (CV). These models, trained on open-domain datasets, excel at extracting and condensing knowledge across multiple domains, exhibiting strong generalization. Following this pathway, recent efforts have fused large-scale, cross-regional spatio-temporal data to develop a unified foundation model~\cite{yuan2024unist,li2024opencity}. However, these spatio-temporal foundation models face substantial challenges in computational demand and memory overhead, primarily arising from their reliance on ``\emph{Transformer + X}" architectures. Specifically, the core self-attention mechanism is utilized to capture temporal dynamics, while the ``X" module, such as GNNs~\cite{li2024opencity,fang2023spatio,wang2025graph}, Graph Transformers, or spatial prompts~\cite{yuan2024unist,dai2024large}, is strategically used to model spatial dependencies. Therefore, given a spatio-temporal graph with $N$ nodes over $T$ time steps, the computational complexity of a fully self-attention model escalates to \emph{unacceptable \textbf{quadratic complexity} of $\mathcal{O}(T^2)$ and $\mathcal{O}(N^2)$.}
As the scale of nodes and the length of time series increase, this complexity bottleneck in transformer-based architectures severely hinders scalability for large-scale traffic graphs and long-term traffic flow series, presenting significant challenges for implementation on edge and low-resource urban devices. 
Therefore, there is an urgent need to develop an efficient model to address the high computational complexities.

% Mamba
Recently, \textbf{Mamba}~\cite{gu2023mamba,qu2024survey}, as a novel state space model (SSM), has gained increasing popularity for its efficient computational capabilities.
Specifically, through introducing selective mechanisms and well-designed hardware-computational algorithms~\cite{qu2024survey}, Mamba delivers comparable modeling capabilities to Transformers while maintaining linear complexity with context length. Furthermore, Mamba-2~\cite{dao2024transformers} extends these capabilities by proposing matrix multiplication based on the Structured State Space Duality (SSD) property, enabling the SSM backbone to perform parallel computations for greater efficiency. Given the advantages of (i) \textit{high performance} facilitated by a selective mechanism, (ii) \textit{rapid training and inference} through matrix multiplication and parallel processing, and (iii) \textit{linear scalability} with context length,
Mamba stands out as an efficient foundation model backbone in various tasks~\cite{qu2024survey}. 
An increasing number of Mamba-based foundation models have been proposed, such as Jamba~\cite{lieber2024jamba} in NLP, Vision Mamba~\cite{zhu2024vision}, and VideoMamba~\cite{li2025videomamba} in CV, Caduceus~\cite{schiff2024caduceus} in genomics, and SSD4Rec~\cite{qu2024ssd4rec} for recommendation. 

Despite Mamba techniques having opened up opportunities to address the high computational complexity limitations of current Transformer-based foundation models, the development of Mamba-based spatio-temporal foundation models, specifically tailored for urban data, remains fully unexplored. 
Directly applying typical Mamba for cross-domain spatio-temporal data suffers from negative transfer and severe performance degradation in zero-shot predictions in unseen regions and cities. 
This limited generalization capability can be attributed to the following key factors:
(1) \textbf{Spatio-temporal Heterogeneity}. Unlike language data, which typically share a common vocabulary, urban data lack explicit cross-domain invariants necessary for effective generalization due to inherent spatio-temporal heterogeneity, as illustrated in Fig.~\ref{fig:motivation}. Temporally, differences in traffic volumes, periods, trends, and timescales lead to inconsistent dynamics across domains. Spatially, variations in city layouts, road networks, and the distribution of transit hubs create further heterogeneity. Such spatio-temporal heterogeneity leads to significant discrepancies in cross-domain urban data distributions~\cite{ji2023spatio,li2024opencity,liu2024one}.
(2) \textbf{Lack of Cross-Domain Adaptation}. The Mamba was originally designed for language sequences and lacks the necessary mechanisms to handle distributional discrepancies across domains, which are prevalent in urban data~\cite{li2024opencity}. As a result, it struggles to adapt to the heterogeneous patterns inherent in cross-domain urban scenarios.
(3) \textbf{Recursive Drift Limits Generalization}. The recursive mechanism of hidden states $h(t)$ and the state transition matrix $\mathbf{A}$ in Mamba effectively captures and propagates past patterns. However, this recursive mechanism may lead to negative effects in the presence of data distribution change~\cite{long2024dgmamba}. Specifically, when training on data from multiple domains, domain-specific features can be inadvertently accumulated or even amplified in the hidden states. This amplification may optimize performance on seen domains but ultimately hinders the model’s ability to generalize to unseen domains.

Motivated by the above observations, we aim to address the inherent heterogeneity challenge and improve the generalization capability of the Mamba backbone through three key steps: (1) capturing domain-specific discriminative features, (2) identifying and aligning cross-domain commonalities, and (3) balancing the fine line between domain-specific discrimination and shared commonalities. Toward this goal, in contrast to existing urban foundation models that optimize a shared representation space through cross-domain data fusion training to fit all datasets, we propose to partition the Mamba representation space into two components, as illustrated in Fig.~\ref{fig:subspace}: a shared subspace for learning cross-domain common patterns $C$, and distinct, non-shared subspaces for domain-specific spatio-temporal patterns $\{S_1, S_2, S_3\}$.

To this end, we propose a \textbf{D}omain \textbf{a}daptive Ma\textbf{mba} for efficient urban \textbf{S}patio-\textbf{T}emporal prediction, i.e., \textbf{Damba-ST}. It excels in strong generalizability across domains while maintaining the Mamba advantages of linear complexity. Damba-ST comprises three key modules: \textit{Multi-View Encoding (MVE)}, \textit{Intra-Domain Scanning (IDS)}, and \textit{Cross-Domain Adaptation (CDA)}. 
Specifically, the MVE module decomposes the spatio-temporal data into three distinct views to effectively capture spatial, temporal, and spatio-temporal delay dependencies, and introduces ``Domain-Adapters" that function as bridges to facilitate cross-domain pattern communication. The IDS module employs three scanning strategies to standardize the data from the three perspectives into a uniform sequence format. Then, these standardized sequences are fed into the CDA module to learn domain-adaptive representations. The CDA module consists of three variants of the \textit{Domain-Adaptive State Space Model (DASSM)}, each comprising a Discrimination Learner, an Adapter Learner, and a Commonalities Learner. The Discrimination Learner captures domain-specific discriminative features, the Adapter Learner serves as a bridge to facilitate information exchange, and the Commonalities Learner identifies and aligns underlying patterns shared across domains. Finally, a fusion module combines features from three views and predicts future values. To sum up, our major contributions include:

\begin{itemize}[leftmargin=*]
    \item We introduce Damba-ST, a novel spatio-temporal prediction backbone with efficient linear computational complexity. To the best of our knowledge, Damba-ST is the first urban model designed to efficiently tackle the challenges of heterogeneity and quadratic computational complexity while exploring the generalizability of State Space Models in spatio-temporal contexts.
    \item We propose the Domain-Adaptive State Space Model, which addresses the heterogeneity challenge by partitioning the representation space into a shared subspace for learning cross-domain commonalities and independent subspaces for capturing domain-specific discriminative features. 
    Additionally, we introduce the Domain Adapters that serve as domain-aware proxies to bridge disparate domain data and facilitate the alignment of cross-domain commonalities. 
    \item Extensive experiments demonstrate the generalization and efficiency of Damba-ST. It achieves state-of-the-art performance on prediction tasks and demonstrates strong zero-shot generalization, enabling seamless deployment in new traffic environments without extensive retraining or fine-tuning.
\end{itemize}

% \revise{The remainder of this paper is organized as follows. Section~\ref{sec:pre} formalizes the problem and reviews Mamba preliminaries. Section~\ref{sec:method} introduces the proposed Damba-ST framework. Section~\ref{sec:optimize} presents the model optimization and theoretical analysis. Section~\ref{sec:exp} details the experimental setup and reports the results. Section~\ref{sec:literature} surveys recent advances in spatio-temporal foundation models and Mamba-based spatio-temporal models. Section~\ref{sec:conclusion} concludes the paper.}

\section{Preliminaries}
\label{sec:pre}

\subsection{Problem Formulation}

\subsubsection{\textbf{Urban Spatio-Temporal Data}}
Urban flow is inherently spatio-temporal, reflecting the spatial distribution and temporal dynamics of a city.
\begin{itemize}
    \item The \textit{spatial component} is structured as a graph $\mathcal{G}=(\mathcal{V},\mathcal{E},\mathcal{A})$, where the node set $\mathcal{V}=\{v_1,\dots,v_N\}$ represents the $N$ sensors, the edge set $\mathcal{E}$ defines connectivity among sensors, and $\mathcal{A}\in\mathbb{R}^{N\times N}$ is a weighted adjacency matrix encoding pairwise spatial proximity~\cite{li2018dcrnn_traffic,ShaoZWWXCJ22}.

    \item The \textit{temporal component} $\mathcal{X}=(\mathcal{X}_1,\dots,\mathcal{X}_T)\in\mathbb{R}^{T\times N\times C_{\text{in}}}$ represents a multivariate time series over $T$ time steps for $N$ sensors, where $\mathcal{X}_t\in\mathbb{R}^{N\times C_{\text{in}}}$ stacks per-sensor channels (e.g., target metric, time-of-day, day-of-week). We denote $\mathcal{X}_{t_a:t_b}\in\mathbb{R}^{(t_b-t_a+1)\times N\times C_{\text{in}}}$ for a temporal slice, and $\mathcal{X}_{:,n}\in\mathbb{R}^{T\times C_{\text{in}}}$ for the time series at sensor $n$ over $T$ time steps.

    \item The \textit{pre-training data} consists of the spatial and temporal components from $M$ cities, denoted by $\mathbf{D}_{\mathrm{train}}=\{(\mathcal{X}^i,\mathcal{G}^i)\}_{i=1}^{M}$.
\end{itemize}

\subsubsection{\textbf{Urban Spatio-Temporal Prediction}}
At any reference time index $t$, given the past $H$ steps
$\mathcal{X}_{t-H+1:t}^i\in\mathbb{R}^{H\times N\times C_{\text{in}}}$
and the spatial graph $\mathcal{G}^i$ of city $i$, the goal is to predict the next $F$ steps.
We define the supervision target as
$\mathcal{Y}_{t+1:t+F}^i \in \mathbb{R}^{F\times N\times C_{\text{out}}}$,
and train a spatio-temporal foundation model $\mathcal{F}_\Theta(\cdot)$:
\begin{equation}
\label{eq:Preliminarie1}
\hat{\mathcal{Y}}_{t+1:t+F}^i
= \mathcal{F}_\Theta\!\big(\mathcal{X}_{t-H+1:t}^i,\ \mathcal{G}^i\big).
\end{equation}

\subsubsection{\textbf{Spatio-Temporal Zero-Shot Generalization}}
The pretrained model is directly applied to unseen urban datasets
$\mathbf{D}_{\mathrm{unseen}}=\{(\mathcal{X}^j,\mathcal{G}^j)\mid j>M\}$ without any retraining or fine-tuning.
For each unseen city $j$, we similarly define the ground-truth target
$\mathcal{Y}_{t+1:t+F}^j \in \mathbb{R}^{F\times N\times C_{\text{out}}}$ as the future values of the main traffic metric
extracted from $\mathcal{X}^j$. Zero-shot generalization is then defined as:
\begin{align}
\label{eq:Generalization}
\hat{\mathcal{Y}}_{t+1:t+F}^{j}
= \mathcal{F}_{\Theta_{\mathrm{frozen}}}\!\big(\mathcal{X}_{t-H+1:t}^{j},\ \mathcal{G}^{j}\big).
\end{align}

\subsection{Mamba}
\subsubsection{\textbf{State Space Model (SSM)}}The classical State Space Models (SSMs) describes the state evolution of a linear time-invariant system, which map input signal $\mathbf{x}=\{x(1),\cdots,x(t),\cdots\} \in \mathbb{R}^{L\times D} \mapsto \mathbf{y}=\{y(1),\cdots,y(t),\cdots\} \in \mathbb{R}^{L \times D}$ through implicit latent state $h(t) \in \mathbb{R}^{N\times D}$, where $t$, $L$, $D$ and $N$ indicate the time step, sequence length, channel number of the signal and state size, respectively.
These models can be formulated as the following linear ordinary differential equations:
\begin{equation}
\label{eq:ode}
\begin{split}
h^{\prime}(t)=\mathbf{A}h(t)+\mathbf{B}x(t),\quad
y(t)=\mathbf{C}h(t)+\mathbf{D}x(t),
\end{split}
\end{equation}
where $\mathbf{A} \in \mathbb{R}^{N\times N} $ is the state transition matrix that describes how states change over time, $\mathbf{B} \in \mathbb{R}^{N\times D} $ is the input matrix that controls how inputs affect state changes, $\mathbf{C} \in \mathbb{R}^{N\times D} $ denotes the output matrix that indicates how outputs are generated based on current states and $\mathbf{D} \in \mathbb{R}^{D} $ represents the command coefficient that determines how inputs affect outputs directly. Most SSMs exclude the second term in the observation equation, i.e., set $\mathbf{D}x(t)=0$, which can be recognized as a skip connection in deep learning models.
The time-continuous nature poses challenges for integration into deep learning architectures.
To alleviate this issue, most methods utilize the Zero-Order Hold rule~\cite{gu2023mamba} to discretize continuous time into $K$ intervals, which assumes that the function value remains constant over the interval $\boldsymbol{\Delta} \in \mathbb{R}^D$. The Eq.~\eqref{eq:ode} can be reformulated as:
\begin{equation}
\label{eq:Discrete}
\begin{split}
    h_t = \overline{\mathbf{A}}  h_{t-1} + \overline{\mathbf{B}}  x_t,
    \quad y_t = \mathbf{C}  h_t, 
\end{split}
\end{equation}
where $\overline{\mathbf{A}} = \exp\left( \boldsymbol{\Delta} \mathbf{A} \right)$ and  $\overline{\mathbf{B}}=(\boldsymbol{\Delta} \mathbf{A})^{-1}(\exp (\boldsymbol{\Delta} \mathbf{A})-\mathbf{I}) \cdot \boldsymbol{\Delta} \mathbf{B}$.
Discrete SSMs can be interpreted as a combination of CNNs and RNNs. Typically, the model employs a convolutional mode for efficient, parallelizable training and switches to a recurrent mode for efficient autoregressive inference. The formulations in Eq.~\eqref{eq:Discrete} are equivalent to the following convolution~\cite{gu2020hippo}:
\begin{equation}
\begin{split}
&\overline{\mathbf{K}} = (\mathbf{C} \overline{\mathbf{B}},\mathbf{C} \overline{\mathbf{A}}\overline{\mathbf{B}},\cdots,\mathbf{C} \overline{\mathbf{A}}^k \overline{\mathbf{B}}, \cdots),\\
&\mathbf{y} = \mathbf{x} * \overline{\mathbf{K}},    
\end{split}
\end{equation}
Thus, the overall process can be represented as:
\begin{equation}
\mathbf{y} = \textbf{SSM}(\overline{\mathbf{A}}, \overline{\mathbf{B}}, \mathbf{C})(\mathbf{x}).
\end{equation}
% \yf{Eq.~(2) includes a $D$ term. However, the final expression omits this term without explanation. It would be helpful to clarify.}

\subsubsection{\textbf{Selective State Space Model (Mamba)}}
The discrete SSMs are based on data-independent parameters, meaning that parameters $\bar{\mathbf{A}}$, $\bar{\mathbf{B}}$, and ${\mathbf{C}}$ are time-invariant and the same for any input, limiting their effectiveness in compressing context into a smaller state~\cite{gu2023mamba}. 
Mamba~\cite{gu2023mamba} introduces a selective mechanism to filter out extraneous information while retaining pertinent details indefinitely.
Specifically, it utilizes Linear Projection to parameterize the weight matrices $\{\mathbf{B}_{t}\}_{t=1}^{L} $, $\{\mathbf{C}_{t}\}_{t=1}^{L} $ and $\{\mathbf{\Delta}_{t}\}_{t=1}^{L}$ according to model input $\{x_t\}_{t=1}^{L}$, improving the context-aware ability, i.e.,:
\begin{equation}
\begin{split}
    &\overline{\mathbf{B}}_t = \texttt{Linear}_{\mathbf{B}}(x_t),\\
    &\overline{\mathbf{C}}_t = \texttt{Linear}_{\mathbf{C}}(x_t), \\
    &\boldsymbol{\Delta}_t = \text{Softplus}\left(\texttt{Linear}_{\boldsymbol{\Delta}}(x_t)\right).
\end{split}
\end{equation}
Then the output sequence $\{y_{t}\}_{i=t}^{L} $ can be computed with those input-adaptive discretized parameters as follows:
\begin{equation}
\label{eq:ssm}
h_{t}=\overline{\mathbf{A}}_{t} h_{t-1}+\overline{\mathbf{B}}_{t} x_{t}, \quad y_{t}=\overline{\mathbf{C}_{t}} h_{t}.
\end{equation}
The selection mechanism hinders SSMs from supporting parallel training like CNNs. To overcome this, Mamba adopts Parallel Associative Scan~\cite{harris2007parallel} and Memory Recomputation. The former exploits linear associativity and the parallelism of GPUs and TPUs for memory-efficient computation, while the latter reduces memory usage by recomputing intermediate states during backpropagation.

\renewcommand*{\thefootnote}{\arabic{footnote}}
\section{Damba-ST}
\label{sec:method}

\begin{figure*}
\centerline{\includegraphics[width=0.98\textwidth]{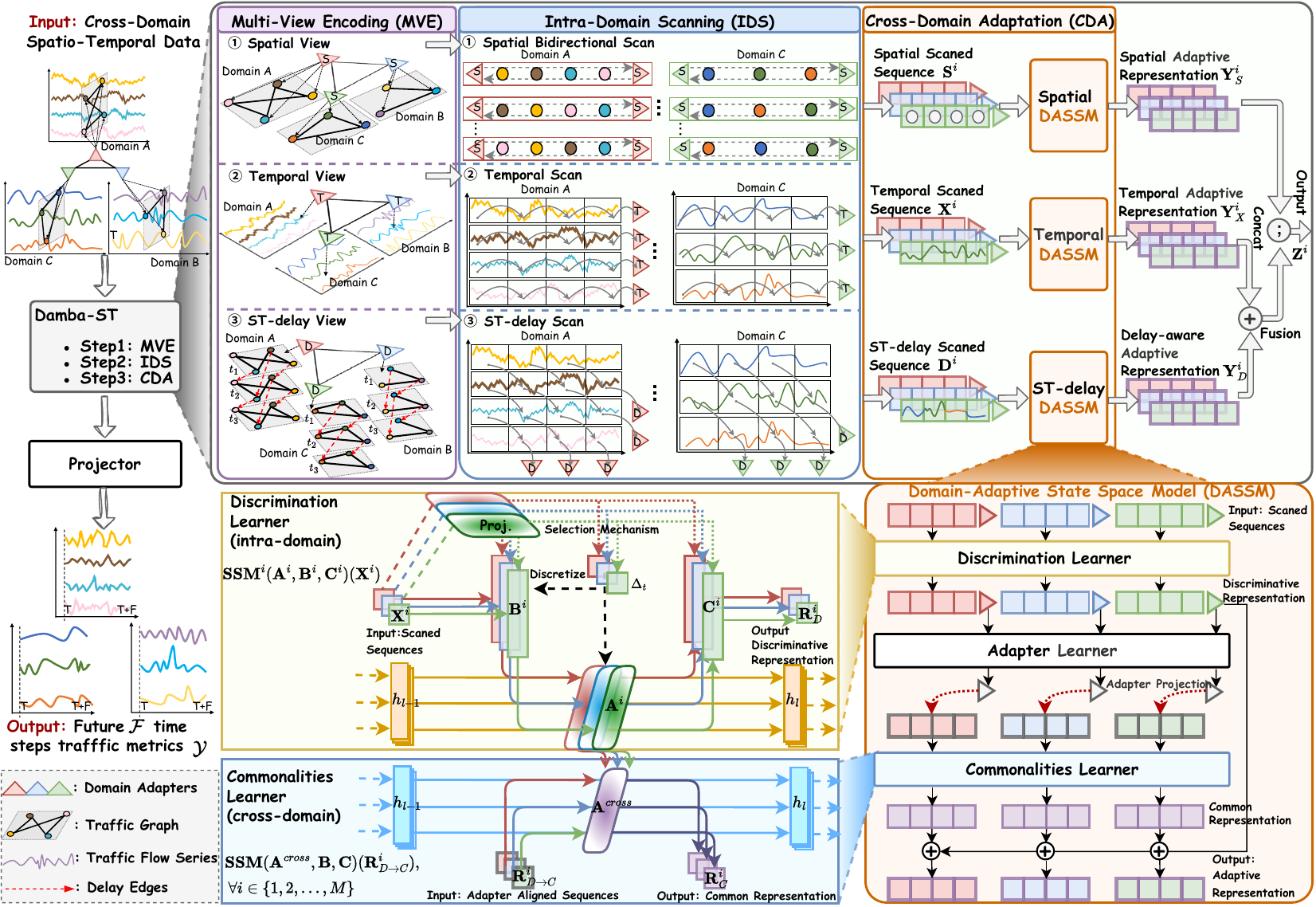}}
\caption{\textbf{Overall framework.} Damba-ST comprises three key modules: \textit{Multi-View Encoding (MVE)}, \textit{Intra-Domain Scanning (IDS)}, and \textit{Cross-Domain Adaptation (CDA)}. The CDA module integrates three variants of the \textit{Domain-Adaptive State Space Model (DASSM)}: Spatial, Temporal, and ST-Delay DASSM. Each variant contains a \textit{Discrimination Learner}, an \textit{Adapter Learner}, and a \textit{Commonalities Learner}.}
\label{fig:framework}
% \vskip -0.1in
\end{figure*}

In this section, we first introduce the novel concept of \textit{Domain Adapters}, followed by a structural overview of the proposed Damba-ST. Subsequently, we present a detailed explanation of each model component.

\subsection{Domain Adapters} 
Although spatio-temporal data lacks explicit cross-domain invariants for effective training and generalization, universal spatio-temporal principles are implicitly shared across diverse scenarios. For instance, while city layouts vary significantly, the relationships between different urban functional areas exhibit common structural patterns. Likewise, different cities may exhibit distinct temporal profiles, but they consistently follow comparable recurrent patterns, including daily cycles, weekly routines, and seasonal or holiday effects. These recurring structures constitute universal temporal regularities shared across cities.

To ensure that each domain retains its unique spatio-temporal characteristics while sharing these fundamental properties within a broader interconnected system, we introduce \textit{Domain Adapters} tailored to each individual domain dataset. These adapters are randomly initialized as learnable embeddings that capture intra-domain underlying patterns. Furthermore, they are interconnected to serve as domain-specific proxies that communicate with one another, facilitating information exchange and bridging previously disconnected domain data. This mechanism is similar to the \verb|[CLS]| token in ViT~\cite{dosovitskiy2020image} and BERT, where a “blank slate” token as the sole input forces the model to encode a “general representation” of the entire sequence. Similar observations can be found in ~\cite{zou2024closer, darcet2023vision}, which demonstrate that the \verb|[CLS]| token naturally absorbs domain-specific information, representing generalizable low-level information. 

\subsection{Structure Overview}

As shown in Fig.~\ref{fig:framework}, Damba-ST processes cross-domain spatio-temporal data through three core modules: \textit{Multi-View Encoding (MVE)}, \textit{Intra-Domain Scanning (IDS)}, and \textit{Cross-Domain Adaptation (CDA)}.
The MVE module decomposes the data into spatial, temporal, and spatio-temporal delay views. The IDS module then standardizes these views into a unified sequential format. The CDA module integrates three variants of the \textit{Domain-Adaptive State Space Model (DASSM)}: Spatial DASSM, Temporal DASSM, and ST-delay DASSM. Within each DASSM variant, the \textit{Discrimination Learner} captures domain-specific patterns, the \textit{Adapter Learner} promotes cross-domain generalization, and the \textit{Commonalities Learner} extracts shared patterns. Finally, a fusion module combines the domain-adaptive representations from the three DASSM variants, and a linear projection layer predicts the traffic metric for the next $F$ time steps.

\subsection{Multi-View Encoding (MVE)}
To better understand the multifaceted factors influencing urban dynamic patterns, we decompose the spatio-temporal (ST) data into three views: Spatial, Temporal, and ST-delay, and introduce three distinct adapters to capture their respective dependencies.
The encoding process for each view consists of two steps: intra-domain feature initialization and adapter initialization.

\subsubsection{\textbf{Spatial View Encoding}} 
Traffic patterns vary across regions due to their unique geographical characteristics. For instance, sites with high connectivity typically experience higher traffic volumes.
To capture the structural information of the traffic graph, we initialize node embeddings using topology-related features. Specifically, we compute the normalized Laplacian matrix of the traffic graph $\mathcal{G}^i$, defined as $\mathbf{\triangle} = \mathbf{I} - \mathbf{D}^{-1/2}\mathcal{A}\mathbf{D}^{-1/2}$, where $\mathbf{I}$ and $\mathbf{D}$ denote the identity and degree matrices of $\mathcal{G}^i$, respectively.
We then perform eigendecomposition of the Laplacian matrix, yielding $\mathbf{\triangle} = \mathbf{U} \mathbf{\Lambda} \mathbf{U}^\mathsf{T}$, where $\mathbf{U}$ and $\mathbf{\Lambda}$ represent the eigenvector and eigenvalue matrices. The $k$ smallest non-trivial eigenvectors are selected to form the initial node encodings, denoted as $\Phi \in \mathbb{R}^{N \times k}$.
These encodings are subsequently projected through a linear layer to obtain the spatial representations $\mathbf{S}^i = \{\mathbf{s}_1, \mathbf{s}_2, \cdots, \mathbf{s}_N\} \in \mathbb{R}^{N \times D}$ for the $N$ nodes in traffic graph $\mathcal{G}^i$.

The \textit{spatio-adapter} $\mathcal{S}^i$ of city $i$ is a virtual node fully connected to all other nodes in the graph $\mathcal{G}^i$. It is initialized as a learnable embedding $\mathcal{S}^i \in \mathbb{R}^D$. This flexible design allows the spatio-adapter to iteratively refine its representation during training, enabling it to effectively capture intra-domain global spatial dependencies.

\subsubsection{\textbf{Temporal View Encoding}} 
For the temporal component, we apply Reverse Instance Normalization~\cite{kim2021reversible} to each individual time series at every time step to alleviate temporal distribution shifts.
Subsequently, to reduce time-step redundancy and extract local temporal semantics~\cite{PatchTST}, we tokenize the normalized time series 
$\mathcal{X}^i=\{\mathcal{X}_1,\dots,\mathcal{X}_T\}\in\mathbb{R}^{T\times N\times C_{\text{in}}}$ into discrete patches using a 1D convolutional layer with output channel dimension $D$, kernel size $P$, and stride $S$.
The number of patches is computed as $L = \left\lfloor \frac{T - P}{S} \right\rfloor + 1$, yielding the patch-level representation $\mathbf{X}^i = \{\mathbf{x}_1, \mathbf{x}_2, \cdots, \mathbf{x}_L\} \in \mathbb{R}^{L \times N \times D}$, where the $n$-th time series is tokenized as $X_{:,n} = \{x_{1,n}, x_{2,n}, \cdots, x_{L,n}\} \in \mathbb{R}^{L \times D}$.

The \textit{temporal-adapter} $\mathcal{T}^i \in \mathbb{R}^D$ for city $i$ is a learnable patch embedding that captures intra-domain latent temporal transition dynamics through iterative updates during model training.

\subsubsection{\textbf{ST-delay View Encoding}} 
There is a critical yet often underestimated dependency in urban modeling: spatio-temporal delay. In real-world traffic scenarios, delay propagation commonly exists among neighboring nodes. For instance, when a traffic incident occurs in a particular region, it may take several minutes to affect traffic conditions in adjacent areas. However, this delay effect is largely overlooked in existing traffic forecasting models.
To accurately capture propagation delays, we propose a two-step strategy. Specifically, for each connected node pair $(v_a, v_b)$, the propagation delay $\tau_{ab}$ is estimated by maximizing the cross-correlation~\cite{azaria1984time} between their respective time series $\mathcal{X}_{:,a}\in\mathbb{R}^{T\times C_{\text{in}}}$ and $\mathcal{X}_{:,b}\in\mathbb{R}^{T\times C_{\text{in}}}$ via temporal shifting~\cite{long2024unveiling}, formulated as:
\begin{equation}
\tau_{ab} = \arg \max_t \textit{corr}(\mathcal{X}_{:,a} \overset{t}{\rightarrow} \mathcal{X}_{:,b}), 
\label{mcc}
\end{equation}
where $\overset{t}{\rightarrow}$ denotes a $t$-step shift applied to $\mathcal{X}_{:,a}$, and $\textit{corr}(\cdot)$ represents the Pearson correlation function.
We estimate the delay values for all connected node pairs in advance and obtain the delay propagation adjacency matrix $\bm{\tau} \in \mathbb{R}^{N \times N}$, which encodes the delay value between connected node pairs.
Additionally, since traffic delays are dynamically influenced by temporal factors, such as longer delays during commuting hours, we use the Multilayer Perceptron (MLP) to transform timestamps as timestamp-aware features $\bm{\tau}(t)=\text{MLP}(t)$ to adjust the base delays dynamically. Here, the timestamp feature is defined as ${t} = \{t^d; t^w\} \in \mathbb{R}^{T \times 2}$, where $t^d \in \mathbb{R}^T$ represents the time of day and $t^w \in \mathbb{R}^T$ denotes the day of the week. 
The final delay matrix in city $i$ is computed as $\bm{\tau}^i = \bm{\tau} + \bm{\tau}(t)$, which serves as an indicator for scanning relevant historical temporal patches from neighboring nodes in the Section~\ref{ST Delay Scan} \textbf{\textit{ST-delay Scan}}.

The \textit{ST-delay adapter} $\mathcal{D}^i$ for city $i$ is prepended with a learnable embedding designed to capture latent delay dependencies.

\subsection{Intra-Domain Scanning (IDS)}
In this subsection, we introduce three scanning strategies to standardize the three views of the \textit{MVE} module into a uniform sequential format for subsequent learning.

\subsubsection{\textbf{Spatial Bidirectional Scan}}
\label{Spatial Bidirectional Scan}
Given that topological nodes do not have a specific order, we employ a bidirectional scan to transform the spatial structure into a uniform sequential format. Specifically, we perform random walks on the traffic graph to generate $N$ paths, each starting from a distinct node $v_a$ with length $L-2$, such as $\text{path}_{v_a} = (v_a \rightarrow v_b \rightarrow \cdots \rightarrow v_g)$.
To enable the \textit{spatio-adapter} $\mathcal{S}^i$ to capture global spatial dependencies, we attach it to both the beginning and the end of each path.
Since spatial nodes are unordered, we scan each path in both forward and backward directions to eliminate sequential order bias. For example, the standardized spatial sequences of $\text{path}_{v_a}$ are:
\begin{equation}
\begin{split}
\label{eq: Bidirectional Scan}
    &\overset{\rightarrow}{\mathbf{S}}_{v_a}=\{\mathcal{S}^i; \  \mathbf{s}_a,\mathbf{s}_b,\cdots,\mathbf{s}_g; \   \mathcal{S}^i\} \in \mathbb{R}^{L \times D},\\
    &\overset{\leftarrow}{\mathbf{S}}_{v_a}=\{\mathcal{S}^i; \  \mathbf{s}_g,\cdots,\mathbf{s}_b,\mathbf{s}_{a}; \  \mathcal{S}^i\} \in \mathbb{R}^{L \times D}.
\end{split}
\end{equation}
Then, the standardized spatial sequence for all $N$ paths in city $i$ is expressed as:
\begin{equation}
\label{eq:Spatial scan}
\mathbf{S}^i=\{\overset{\leftrightarrow}{\mathbf{S}}_{v_{1}}, \overset{\leftrightarrow}{\mathbf{S}}_{v_{2}},\cdots, \overset{\leftrightarrow}{\mathbf{S}}_{v_{N}}\} \in \mathbb{R}^{N \times L\times D},
\end{equation}
where $\overset{\leftrightarrow}{\mathbf{S}}_{v_{n}}$ represent the bidirectional spatial sequence of $\text{path}_{v_n}$.

\subsubsection{\textbf{Temporal Scan}} The time series patches naturally follow chronological order. To enable the \textit{temporal-adapter} $\mathcal{T}^i$ to learn underlying temporal state transition patterns, we append $\mathcal{T}^i$ to the end of each series of patches. The standardized temporal sequence series in domain $i$ is:

\begin{equation}
\label{eq:Temporal Scan}
\mathbf{X}^i = \{ \mathbf{x}_1, \mathbf{x}_2, \dots, \mathbf{x}_{L-1}; \  \mathcal{T}^i \} \in \mathbb{R}^{L \times N\times D}.
\end{equation}

\subsubsection{\textbf{ST-delay Scan}} 
\label{ST Delay Scan}
Spatio-temporal delay dependencies propagate along delay edges between graph snapshots. To standardize the ST-delay view into a uniform delay propagation sequence, we connect temporal patches based on the delay matrix $\bm{\tau}^i$.
Specifically, for a connected node path $\text{path}_{v_a}=(v_a \rightarrow v_b \rightarrow v_c)$, a standardized ST-delay propagation sequence begins with the $l$-th temporal patch $x_{l,a}$ of node $v_a$.
According to the delay value $\bm{\tau}_{ab}$, this temporal patch is propagated to the temporal patch $x_{{l + \left\lfloor \frac{\bm{\tau}_{ab}}{P} \right\rfloor},b}$ on node $v_b$ after $\left\lfloor \frac{\bm{\tau}_{ab}}{P} \right\rfloor$ time steps, where $P$ denotes the patch length.
Similarly, the propagation continues to node $v_c$ after an additional $\left\lfloor \frac{\bm{\tau}_{bc}}{P} \right\rfloor$ time steps. To capture the latent delay dependencies in each specific city, we append the \textit{ST-delay adapter} $\mathcal{D}^i$ to the end of each scanned delay propagation sequence.
Thus, a complete delay propagation sequence along the node path $\text{path}_{v_a}=(v_a \rightarrow v_b \rightarrow v_c)$ can be expressed as:
\begin{equation}
\label{eq: delay-scan}
\mathbf{D}_{v_a}=\{x_{l,a}, \  x_{{l+\lfloor \frac{\bm{\tau}_{ab}}{P} \rfloor},b}, \  x_{{l+\lfloor \frac{\bm{\tau}_{ab}}{P} \rfloor+\lfloor \frac{\bm{\tau}_{bc}}{P} \rfloor},c}; \ \mathcal{D}^i \}. 
\end{equation}
We fix the number of patches in each delay path to $L$, then, the standardized delay propagation sequence for all $N$ paths in domain $i$ is expressed as:
\begin{equation}
\label{eq:Delay Scan}
\mathbf{D}^i=\{\mathbf{D}_{v_1}, \mathbf{D}_{v_2}, \cdots, \mathbf{D}_{v_N}\}\in \mathbb{R}^{N \times L\times D}.
\end{equation}
\subsection{Cross-Domain Adaptation (CDA)}
The CDA module integrates three variants of the \textit{Domain-Adaptive State Space Model (DASSM)}: Spatial DASSM, Temporal DASSM, and ST-delay DASSM, each designed to learn domain-adaptive representations.

In the remainder of this section, we use the temporal scanned sequences $\mathbf{X}^i$ as an illustrative example to detail the design of the proposed DASSM. As shown in Fig.~\ref{fig:framework}, DASSM comprises three primary components:
1) the \textit{Discrimination Learner}, which captures domain-specific patterns through intra-domain optimization;
2) the \textit{Adapter Learner}, which enables cross-domain generalization via adapter communication; and
3) the \textit{Commonalities Learner}, which extracts shared patterns through cross-domain fusion optimization.

\subsubsection{\textbf{Discrimination Learner}}

In contrast to existing foundation models~\cite{yuan2024unist,li2024opencity} that fuse cross-domain data to learn a shared representation space, we propose the Discrimination Learner, which optimizes independent, domain-specific subspaces for each domain. This design enables the model to fully exploit domain-specific patterns induced by inherent heterogeneity.
Given the standardized temporal sequence $\mathbf{X}^i = \{ \mathbf{x}_1, \mathbf{x}_2, \dots, \mathbf{x}_{L-1}; \  \mathcal{T}^i \} \in \mathbb{R}^{L\times N \times D}$ from Eq.~\eqref{eq:Temporal Scan}, where $i \in \{1, 2, \dots, M\}$ is the domain index, the Discrimination Learner trains a selective state space model $\textbf{SSM}^i$ for each domain $i$ to capture intra-domain token dependencies. This process can be formulated as:
\begin{equation}
\label{eq:discriminative representation}
\mathbf{R}_D^i = \textbf{SSM}^i(\mathbf{A}^i, \mathbf{B}^i, \mathbf{C}^i)(\mathbf{X}^i),
\end{equation}
where $\mathbf{R}_D^i\in \mathbb{R}^{L \times N\times D}$ is discriminative representation of sequence $\mathbf{X}^i$. 

The domain-adapter $\mathcal{T}^i \in \mathbb{R}^{D}$ is appended as the final token in the sequence and possesses a global receptive field, enabling it to effectively capture and aggregate global dependency information. Its learnable embedding is continuously refined during training. The computation process is defined as follows:
\begin{equation}
\label{eq:adapter_learn}
\begin{split}
\mathbf{h}^i_{L} &= \mathbf{A}_{L}^i \mathbf{h}_{L-1} + \mathbf{B}_{L}^i \mathcal{T}^i, \\
\mathcal{T}^i &= \mathbf{C}_{L}^i\mathbf{h}^i_{L}.
\end{split}
\end{equation}

\subsubsection{\textbf{Adapter Learner}} 
Similar to the \verb|[CLS]| token in ViT~\cite{dosovitskiy2020image}, the domain-adapter $\mathcal{T}^i$ serves as a proxy of intra-domain latent patterns.
To facilitate cross-domain knowledge transfer, we introduce the Adapter Learner, which enables communication among domain adapters and promotes high-level knowledge sharing across domains.
Specifically, the domain-adapters from all $M$ domains are concatenated to form an adapter sequence $\mathcal{T}= \{\mathcal{T}^1, \mathcal{T}^2, \cdots, \mathcal{T}^M\} \in \mathbb{R}^{M\times D}$.
We then randomly shuffle the sequence order and apply both forward and backward SSM to ensure efficient bidirectional information exchange among adapters. This process can be formulated as:
\begin{equation}
\label{eq:adapters sequence}
\hat{\mathcal{T}} = \overset{\rightarrow}{\textbf{SSM}}(\mathbf{A}, \mathbf{B}, \mathbf{C})(\mathcal{T})+\overset{\leftarrow}{\textbf{SSM}}(\mathbf{A}, \mathbf{B}, \mathbf{C})(\mathcal{T}),
\end{equation}
where $ \hat{\mathcal{T}} =\{\hat{\mathcal{T}}^1, \hat{\mathcal{T}}^2,\cdots,\hat{\mathcal{T}}^M\} \in \mathbb{R}^{M \times D} $ denotes the set of domain-adapters after exchanging information within a larger, interconnected system through adapter communication.
Thus, $\hat{\mathcal{T}}$ can be interpreted as the common pattern shared across domains.

\subsubsection{\textbf{Commonalities Learner}} 
To optimize the shared representation subspace, we propose the Commonalities Learner, which harmonizes heterogeneous domain data and aligns cross-domain commonalities through intra-domain feature projection and cross-domain fusion optimization.
Specifically, we first mitigate cross-domain heterogeneity by projecting the intra-domain discriminative representations $\mathbf{R}_D^i$ onto their corresponding domain-adapter $\hat{\mathcal{T}}^i$. This projection is defined as:
\begin{equation}
\label{eq:proj}
\mathbf{R}_{D\rightarrow C}^i =\text{proj}(\mathbf{R}_D^i,\hat{\mathcal{T}}^i)= \left( \frac{\mathbf{R}_D^i \hat{\mathcal{T}}^{i\top}}{\|\hat{\mathcal{T}}^i\|^2} \right) \hat{\mathcal{T}}^i.
\end{equation}
This projection operation can be seen as adapter-induced re-embedding, which eliminates domain-specific biases and maps representations into the \textit{shared (domain-invariant) subspace} captured by the domain adapter $\hat{\mathcal{T}}$. 

Then, we optimize the shared subspace by cross-domain fusion training:
\begin{equation}
\label{eq:commonalities}
\mathbf{R}_C^i = \textbf{SSM}(\mathbf{A}^{\text{cross}}, \mathbf{B}, \mathbf{C})(\mathbf{R}_{D\rightarrow C}^i),   \quad \forall i \in \{1, 2, \ldots, M\},
\end{equation}
where the shared state transition matrix $\mathbf{A}^{\text{cross}}$ is initialized as the mean of the matrices $\mathbf{A}^i$ learned by the $M$ Discrimination Learners, i.e., $\mathbf{A}^{\text{cross}} = \frac{1}{M} \sum_{i=1}^{M} \mathbf{A}^i$.
The resulting $\mathbf{R}_C^i \in \mathbb{R}^{L\times N \times D}$ represents the cross-domain common representation.

Finally, we fuse the Discriminative Representation $\mathbf{R}_D^i$ and the Common Representation $\mathbf{R}_C^i$ using a fusion layer to obtain the final domain-adaptive representation $\mathbf{Y}^i$ for domain $i$, defined as:
\begin{equation}
\label{eq:fusion}
\mathbf{Y}^i = w_1 \cdot \text{Linear}(\mathbf{R}_D^i) + w_2 \cdot \text{Linear}(\mathbf{R}_C^i),
\end{equation}
where $w_1$ and $w_2$ are hyperparameters that control the relative contributions of the discriminative and common components. 
The overall process of DASSM can be represented as:
\begin{equation}
\mathbf{Y}^i = \text{DASSM}(\mathbf{X}^i), \quad \forall i \in \{1, 2, \ldots, M\}.
\end{equation}

\subsubsection{\textbf{DASSM Variants}}
We adopt DASSM in three specialized variants: $\text{DASSM}_\mathrm{S}$, $\text{DASSM}_\mathrm{T}$, and $\text{DASSM}_\mathrm{D}$.
Specifically, $\text{DASSM}_\mathrm{S}$ captures the spatial adaptive representation $\mathbf{Y}^i_\mathrm{S}$ from the spatial scanned sequences $\mathbf{S}^i$, $\text{DASSM}_\mathrm{T}$ learns the temporal adaptive representation $\mathbf{Y}^i_\mathrm{T}$ from the temporal scanned sequences $\mathbf{X}^i$, and $\text{DASSM}_\mathrm{D}$ extracts the delay-aware spatio-temporal adaptive representation $\mathbf{Y}^i_\mathrm{D}$ from the ST-delay scanned sequences $\mathbf{D}^i$, formulated as:
\begin{equation}
\begin{split}
\mathbf{Y}^i_\mathrm{S} &= \text{DASSM}_\mathrm{S}(\mathbf{S}^i), \\
\mathbf{Y}^i_\mathrm{T} &= \text{DASSM}_\mathrm{T}(\mathbf{X}^i), \\
\mathbf{Y}^i_\mathrm{D} &= \text{DASSM}_\mathrm{D}(\mathbf{D}^i), \quad \forall i \in \{1, 2, \ldots, M\}.
\end{split}
\end{equation}
\subsection{Fusion and Prediction}
We fuse the domain-adaptive representations from the three DASSM variants as follows. First, we form the \textit{temporal component} by element-wise summing the temporal representation $\mathbf{Y}^i_{\mathrm T}$ and the delay-aware representation $\mathbf{Y}^i_{\mathrm D}$. The former captures self-dynamics within each series, while the latter models interaction-driven lagged dependencies; together, they summarize the dynamic temporal behavior. Then, we concatenate this fused temporal component with the time-agnostic spatial representation $\mathbf{Y}^i_{\mathrm S}$, enabling the downstream projector to learn task-specific weights across spatial and temporal channels. The final spatio-temporal adaptive representation is:

\begin{equation}
\label{eq:STfusion}
\mathbf{Z}^i = \left( (\mathbf{Y}^i_\mathrm{T} \oplus \mathbf{Y}^i_\mathrm{D}) ; \mathbf{Y}^i_\mathrm{S} \right).
\end{equation}
After obtaining the final representation from Damba-ST, we apply a linear projection layer to generate the predicted traffic metric for the next $F$ time steps.

\section{Optimization and Theoretical Analysis}
\label{sec:optimize}
\subsection{Optimization}
We adopt the end-to-end supervised training to predict the future $F$ time steps of traffic metrics. The objective function can be formulated as:
\begin{align}
    \label{eq:loss}
    \begin{split}
    &\arg \min_\Theta\sum_{i=1}^M\mathcal{L}^i_\Theta,\\
    \text{s.t.}~\mathcal{L}^i_\Theta &=|(\mathcal{Y}^i ,\mathcal{F}_\Theta(\mathcal{X}^i, \mathcal{G}^i))|,
    \end{split}
\end{align}
where $\Theta$ represents all the learnable parameters in the network, $\mathcal{Y}^i$ is the ground truth. We choose L1 loss as our training objective. Additionally, to enlarge the separation between Discrimination Learner and Commonalities Learner, we introduce two regularization terms in our objective function: 1) Model Differences and 2) Representation Differences~\cite{zhang2017jointly,zhang2023federated}. 
\subsubsection{\textbf{Model Differences Regularization}}
This term constrains the parameter similarity between the Discrimination Learner $\theta_D$ and Commonalities Learner $\theta_C$, which can prevent overfitting and improve generalization. 
We use the well-known exponential kernel to measure the parameter similarity, expressed as:
\begin{equation}
\label{eq:term1}
    S_m(\theta_D, \theta_C) = \exp\left( - \frac{\|\theta_D-\theta_C\|_F}{2\sigma^2} \right),
\end{equation}
where $\|\theta_D-\theta_C\|_F$ is the Frobenius norm, measuring the parameter distance, $\sigma$ is a regularization parameter that controls the range of similarity, and $\exp(\cdot)$ is the exponential function. 
\subsubsection{\textbf{Representation Differences Regularization}}
This term constrains the overlap between discriminative representations $\mathbf{R}_D$ and Common representations $\mathbf{R}_C$. The distance function is defined as:
\begin{equation}
\label{eq:term2}
    S_r(\mathbf{R}_D, \mathbf{R}_C) = \frac{1}{C_0} \|\mathbf{R}_D^T \cdot \mathbf{R}_C\|, 
\end{equation}
where $C_0 > 0$ is a normalization parameter. The final objective function of Urban-SSM is then achieved
by combining Eq.~\eqref{eq:loss} with~\eqref{eq:term1} and~\eqref{eq:term2}:
\begin{align}
    \label{eq:finalloss}
    \begin{split}
    \arg \min_\Theta\sum_{i=1}^M\mathcal{L}^i_\Theta & + \alpha S_m(\theta_D, \theta_C)+ \beta S_r(\mathbf{R}_D, \mathbf{R}_C), 
    \end{split}
\end{align}
where $\alpha$ and $\beta$ are two balance parameters.

\subsection{Domain-Adaptive State Space Mode (DASSM) Analysis}
In this section, we further interpret how domain adapters facilitate cross-domain knowledge transfer in a two-fold manner and provide a theoretical analysis.
\subsubsection{\textbf{Domain Adapters function as Intra-Domain Proxies}} The three domain adapters are randomly initialized as learnable embeddings and appended to the end of each sequence to ensure a global receptive field. The update mechanism consists of two stages:
\begin{itemize}[leftmargin=*]
    \item Intra-Domain Update: During intra-domain training within the Discrimination Learner, each domain adapter aggregates information from all other tokens in the same domain (global feature aggregation), with the computation process defined in Eq.~\eqref{eq:adapter_learn}. In this phase, the SGD backpropagation pushes the Discrimination Learner to encode intra-domain global patterns into the learnable embedding.
    \item Cross-Domain Update: During cross-domain training within the Adapter Learner, the domain adapters from all $M$ domains are concatenated to form an adapter sequence, as defined in Eq.~\eqref{eq:adapters sequence}. Backpropagation then pushes information exchange among adapters, allowing each to encode cross-domain generalizable patterns into its learnable embedding.    
\end{itemize}
\subsubsection{\textbf{The Commonalities Learner Learns a Transformation from Domain-Specific Discriminative Patterns to Generalized Common Patterns}}
\label{sec:theoretical analysis}
It is worth noting that our proposed Commonalities Learner has the same theory support as Graph Prompt~\cite{fang2023universal,zhao2024all}, which has been proven equivalent to any form of prompting function for achieving universal graph transformations.
We extend this idea to the spatio-temporal scenario, enabling the Commonalities Learner to theoretically achieve an effect equivalent to any form of spatio-temporal prompting function. This allows it to transform domain-specific discriminative patterns into generalized common patterns. Specifically, since the domain adapter is a learnable embedding, the projection operation $\text{proj}(\mathbf{R}_D^i,\hat{\mathcal{T}}^i)$ in Eq.~\eqref{eq:proj} is equivalent to adding a learnable prompt $P$ to the original features $\mathbf{R}_D^i$, resulting the prompted features $\mathbf{R}_{D\rightarrow C}^i$, formulated as:
\begin{equation}
\begin{split}
% \mathbf{R}_{D\rightarrow C}^i &=\text{proj}(\mathbf{R}_D^i,\hat{\mathcal{T}}^i)\\
\mathbf{R}_{D\rightarrow C}^i 
&= \left( \frac{\mathbf{R}_D^i \hat{\mathcal{T}}^{i\top}}{\|\hat{\mathcal{T}}^i\|^2} \right) \hat{\mathcal{T}}^i\\
&=\mathbf{R}_D^i+\left( \frac{\mathbf{R}_D^i \hat{\mathcal{T}}^{i\top}}{\|\hat{\mathcal{T}}^i\|^2} \hat{\mathcal{T}}^i-\mathbf{R}_D^i\right) \\
&= \mathbf{R}_D^i + P,
\end{split}
\end{equation}
because domain-adapter $\hat{T}^i$ is a learnable embedding, the residual can serve as adding a learnable prompt $P$ (i.e., correction term) to the intra-domain representation. The Commonalities Learner is trained on these prompted representations to optimize the shared representation subspace. 
Without loss of generality, we symbolize this projection operation as:
\begin{equation}
    \hat{\mathcal{X}}=\mathcal{X}+P
\end{equation}

\begin{theorem}
\label{theorem}
(Generalization Capability of Learnable Prompt $P$)  
Given a Spatio-Temporal model $ \mathcal{F}_\Theta$, and the input spatio-temporal data $(\mathcal{G},\mathcal{X})$.
For any domain generalization transformation function $g \colon \mathbb{D}_{\text{specific}} \rightarrow \mathbb{D}_{\text{common}}$, such that $g(\mathcal{G}, \mathcal{X}) = (\hat{\mathcal{G}}, \hat{\mathcal{X}})$, there exists a learnable prompt $P$ that satisfies:
\begin{align}
    \mathcal{F}_\Theta\big((\mathcal{G}, \mathcal{X}) + P\big) = \mathcal{F}_\Theta\big(g(\mathcal{G}, \mathcal{X})\big).
\end{align}
\end{theorem}
Theorem~\ref{theorem} implies that a learnable prompt $P$ can theoretically induce the same model behavior as applying a domain-generalizing transformation function $g$. Specifically, if $g$ can map domain-specific data $(\mathcal{G}, \mathcal{X})$ into a common representation $(\mathcal{G}^*, \mathcal{X}^*)$ that enables effective generalization, then optimizing the prompt $P$ can likewise enable the model to produce equivalent generalizable outputs. To further illustrate the domain-generalized transformation $g(\cdot)$ into the spatio-temporal scenario, we decompose it into three specific sub-transformations.
\vskip -0.1in
\begin{proposition}
Given the spatial scanned sequences $\mathbf{S}$, temporal scanned sequences $\mathbf{X}$, and ST-delay scanned sequences $\mathbf{D}$. 
Any domain-generalizing transformation function $g\colon\mathbb{D}_{\text{specific}} \rightarrow \mathbb{D}_{\text{common}}$ can be decoupled into the composition of the following sub-transformations:
\begin{itemize}[leftmargin=*]
    \item \textbf{Spatial pattern transformation:} Modifies the spatial node features to obtain the generalized spatial representation $\hat{\mathbf{S}} = g_S(\mathbf{S})$.
    \item \textbf{Temporal pattern transformation:} Modifies the temporal features to obtain the generalized temporal representation $\hat{\mathbf{X}} = g_T(\mathbf{X})$.
    \item \textbf{ST-delay pattern transformation:} Modifies the delay-aware features to obtain the generalized spatio-temporal representation $\hat{\mathbf{D}} = g_D(\mathbf{D})$.
\end{itemize}
\end{proposition}
\textit{Proof.} Please refer to the Supplementary Materials\footnote{https://github.com/RuiAN77/Damba-ST}.

\section{Experiments}
\label{sec:exp}
In this section, we investigate the following key questions to evaluate the effectiveness and efficiency of Damba-ST:
\begin{itemize}[leftmargin=*]
\item \textbf{RQ1}: How does the Damba-ST perform in the In-distribution setting compared to existing spatio-temporal baselines?
\item \textbf{RQ2}: What advantages does our proposed Damba-ST offer over baselines in cross-domain transfer learning?
\item \textbf{RQ3}: What are the individual contributions of the different modules to the performance gains of the Damba-ST?
\item \textbf{RQ4}: Compared to existing spatio-temporal foundation models, what advantages does Damba-ST have in terms of computational efficiency and deployment practicality?
\item \textbf{RQ5}: How do different hyperparameter values affect the performance of Damba-ST?
\end{itemize}
\begin{table*}[!h]
\caption{Results of in-distribution long-term prediction. The best results are highlighted in \boldres{red}, and the second-best results in \secondres{\underline{underlining}}.}
\vskip -0.1in
\label{tab:In-distribution Prediction}
    \renewcommand{\arraystretch}{0.85} 
    \centering
    \resizebox{\textwidth}{!}{
    \begin{threeparttable}
    \begin{small}
    \renewcommand{\multirowsetup}{\centering}
    \setlength{\tabcolsep}{1pt}    
    \begin{tabular}{c c c c c c c c c c c c c c c}
        \toprule
        \multirow{3}{*}{\scalebox{0.8}{Dataset}} & \multirow{3}{*}{\scalebox{0.8}{Metrics}} & 
        \multicolumn{3}{c}{\scalebox{0.8}{Foundation Models}} &\multicolumn{2}{c}{\scalebox{0.8}{Mamba-based Models}} & \multicolumn{3}{c}{\scalebox{0.8}{Attention-based Models}} & \multicolumn{5}{c}{\scalebox{0.8}{GNNs-based Models}} \\
       \cmidrule(lr){3-5}\cmidrule(lr){6-7}\cmidrule(lr){8-10}\cmidrule(lr){11-15}
        & & \scalebox{0.8}{$\textbf{Damba-ST}$} & \scalebox{0.8}{UniST} & \scalebox{0.8}{OpenCity} & \scalebox{0.8}{STG-Mamba} & \scalebox{0.8}{SpoT-Mamba} & \scalebox{0.8}{PDFormer} & \scalebox{0.8}{STWA} & \scalebox{0.8}{ASTGCN} & \scalebox{0.8}{STSGCN} & \scalebox{0.8}{MTGNN} & \scalebox{0.8}{GWN} & \scalebox{0.8}{TGCN} & \scalebox{0.8}{STGCN} \\
        & & \scalebox{0.8}{$\textbf{(Ours)}$} & \scalebox{0.8}{~\cite{yuan2024unist}} & \scalebox{0.8}{~\cite{li2024opencity}} & \scalebox{0.8}{~\cite{li2024stg}} & \scalebox{0.8}{~\cite{choi2024spot}} &  \scalebox{0.8}{~\cite{jiang2023pdformer}} & \scalebox{0.8}{~\cite{cirstea2022towards}} & \scalebox{0.8}{~\cite{guo2019attention}} & \scalebox{0.8}{~\cite{song2020spatial}} & \scalebox{0.8}{~\cite{wu2020connecting}} & \scalebox{0.8}{~\cite{wu2019graph}} & \scalebox{0.8}{~\cite{zhao2019t}} & \scalebox{0.8}{~\cite{yu2018spatio}} \\
        \toprule
        \multirow{3}{*}{\scalebox{0.8}{NYC-TAXI}} 
        & \scalebox{0.8}{MAE} & \scalebox{0.8}{\boldres{2.89}} & \scalebox{0.8}{3.52} & \scalebox{0.8}{\secondres{3.10}} & \scalebox{0.8}{4.56} & \scalebox{0.8}{4.13} & \scalebox{0.8}{3.64} & \scalebox{0.8}{6.05} & \scalebox{0.8}{5.45} & \scalebox{0.8}{5.03} & \scalebox{0.8}{3.72} & \scalebox{0.8}{4.28} & \scalebox{0.8}{6.10} & \scalebox{0.8}{4.17} \\
        
        & \scalebox{0.8}{RMSE} & \scalebox{0.8}{\boldres{6.23}} & \scalebox{0.8}{8.60} & \scalebox{0.8}{\secondres{6.45}} & \scalebox{0.8}{11.46} & \scalebox{0.8}{9.26} & \scalebox{0.8}{8.24} & \scalebox{0.8}{15.22} & \scalebox{0.8}{13.44} & \scalebox{0.8}{10.96} & \scalebox{0.8}{7.94} & \scalebox{0.8}{10.59} & \scalebox{0.8}{12.70} & \scalebox{0.8}{9.19} \\
        
        & \scalebox{0.8}{MAPE} & \scalebox{0.8}{\boldres{36.52\%}} & \scalebox{0.8}{38.29\%} & \scalebox{0.8}{\secondres{37.39\%}} & \scalebox{0.8}{49.92\%} & \scalebox{0.8}{39.42\%}  & \scalebox{0.8}{37.40\%} & \scalebox{0.8}{54.81\%} & \scalebox{0.8}{59.05\%} & \scalebox{0.8}{65.17\%} & \scalebox{0.8}{43.35\%} & \scalebox{0.8}{41.82\%} & \scalebox{0.8}{80.39\%} & \scalebox{0.8}{45.54\%} \\
        \hline
        
        \multirow{3}{*}{\scalebox{0.8}{PEMS-BAY}} 
        & \scalebox{0.8}{MAE} & \scalebox{0.8}{\secondres{2.63}} &-    & \scalebox{0.8}{\boldres{2.59}} & \scalebox{0.8}{2.80} & \scalebox{0.8}{2.77} & \scalebox{0.8}{2.75} & \scalebox{0.8}{2.74} & \scalebox{0.8}{2.82} & \scalebox{0.8}{2.87} & \scalebox{0.8}{2.59} & \scalebox{0.8}{2.66} & \scalebox{0.8}{2.94} & \scalebox{0.8}{2.72} \\
        & \scalebox{0.8}{RMSE} & \scalebox{0.8}{5.71} &-    & \scalebox{0.8}{\secondres{5.67}} & \scalebox{0.8}{6.42} & \scalebox{0.8}{6.82} & \scalebox{0.8}{6.15} & \scalebox{0.8}{5.65} & \scalebox{0.8}{6.31} & \scalebox{0.8}{6.10} & \scalebox{0.8}{5.43} & \scalebox{0.8}{\boldres{5.60}} & \scalebox{0.8}{6.33} & \scalebox{0.8}{5.96} \\
        & \scalebox{0.8}{MAPE} & \scalebox{0.8}{5.97\%} &-    & \scalebox{0.8}{5.99\%} & \scalebox{0.8}{7.28\%} & \scalebox{0.8}{6.23\%} & \scalebox{0.8}{6.50\%} & \scalebox{0.8}{6.03\%} & \scalebox{0.8}{7.06\%} & \scalebox{0.8}{6.80\%} & \scalebox{0.8}{\secondres{5.87\%}} & \scalebox{0.8}{\boldres{5.94\%}} & \scalebox{0.8}{7.23\%} & \scalebox{0.8}{6.55\%} \\
        \hline
        
        \multirow{3}{*}{\scalebox{0.8}{CAD12-2}} 
        & \scalebox{0.8}{MAE} & \scalebox{0.8}{\boldres{22.37}} &-    & \scalebox{0.8}{\secondres{24.20}} & \scalebox{0.8}{36.72} & \scalebox{0.8}{38.22} & \scalebox{0.8}{34.23} & \scalebox{0.8}{41.21} & \scalebox{0.8}{35.19} & \scalebox{0.8}{37.00} & \scalebox{0.8}{36.20} & \scalebox{0.8}{38.05} & \scalebox{0.8}{36.53} & \scalebox{0.8}{34.60} \\
        
        & \scalebox{0.8}{RMSE} & \scalebox{0.8}{\boldres{37.32}} &-    & \scalebox{0.8}{\secondres{39.23}} & \scalebox{0.8}{59.33} & \scalebox{0.8}{70.13} & \scalebox{0.8}{59.70} & \scalebox{0.8}{77.78} & \scalebox{0.8}{57.87} & \scalebox{0.8}{63.19} & \scalebox{0.8}{65.55} & \scalebox{0.8}{69.89} & \scalebox{0.8}{60.01} & \scalebox{0.8}{62.47} \\
        
        & \scalebox{0.8}{MAPE} & \scalebox{0.8}{\boldres{32.58\%}} &-    & \scalebox{0.8}{\secondres{33.22\%}} & \scalebox{0.8}{64.91\%} & \scalebox{0.8}{66.34\%} & \scalebox{0.8}{50.92\%} & \scalebox{0.8}{67.72\%} & \scalebox{0.8}{59.73\%} & \scalebox{0.8}{58.29\%} & \scalebox{0.8}{56.86\%} & \scalebox{0.8}{64.77\%} & \scalebox{0.8}{61.60\%} & \scalebox{0.8}{52.49\%} \\
        \hline
        
        \multirow{3}{*}{\scalebox{0.8}{CAD8-2}} 
        & \scalebox{0.8}{MAE} & \scalebox{0.8}{\secondres{18.35}} &-    & \scalebox{0.8}{\boldres{17.95}} & \scalebox{0.8}{25.91} & \scalebox{0.8}{26.12} & \scalebox{0.8}{24.32} & \scalebox{0.8}{26.57} & \scalebox{0.8}{24.24} & \scalebox{0.8}{24.60} & \scalebox{0.8}{21.85} & \scalebox{0.8}{25.17} & \scalebox{0.8}{23.43} & \scalebox{0.8}{24.30} \\
        
        & \scalebox{0.8}{RMSE} & \scalebox{0.8}{\boldres{29.35}} &-   & \scalebox{0.8}{\secondres{29.61}} & \scalebox{0.8}{39.73} & \scalebox{0.8}{42.77} & \scalebox{0.8}{38.95} & \scalebox{0.8}{44.01} & \scalebox{0.8}{38.14} & \scalebox{0.8}{41.23} & \scalebox{0.8}{34.02} & \scalebox{0.8}{41.47} & \scalebox{0.8}{37.03} & \scalebox{0.8}{39.59} \\
        
        & \scalebox{0.8}{MAPE} & \scalebox{0.8}{\secondres{12.32\%}} &-    & \scalebox{0.8}{\boldres{10.97\%}} & \scalebox{0.8}{19.28\%} & \scalebox{0.8}{19.27\%} & \scalebox{0.8}{14.84\%} & \scalebox{0.8}{20.21\%} & \scalebox{0.8}{18.36\%} & \scalebox{0.8}{18.54\%} & \scalebox{0.8}{14.20\%} & \scalebox{0.8}{18.50\%} & \scalebox{0.8}{15.55\%} & \scalebox{0.8}{18.61\%} \\
        \hline
        
        \multirow{3}{*}{\scalebox{0.8}{CAD8-1}} 
        & \scalebox{0.8}{MAE} & \scalebox{0.8}{\boldres{21.59}} &-    & \scalebox{0.8}{\secondres{22.50}} & \scalebox{0.8}{30.22} & \scalebox{0.8}{30.56} & \scalebox{0.8}{29.64} & \scalebox{0.8}{32.09} & \scalebox{0.8}{29.38} & \scalebox{0.8}{32.38} & \scalebox{0.8}{31.60} & \scalebox{0.8}{29.03} & \scalebox{0.8}{27.68} & \scalebox{0.8}{31.26} \\
        
        & \scalebox{0.8}{RMSE} & \scalebox{0.8}{\boldres{38.35}} &-    & \scalebox{0.8}{\secondres{39.06}} & \scalebox{0.8}{49.98} & \scalebox{0.8}{51.34} & \scalebox{0.8}{49.79} & \scalebox{0.8}{32.09} & \scalebox{0.8}{46.84} & \scalebox{0.8}{53.28} & \scalebox{0.8}{53.01} & \scalebox{0.8}{49.26} & \scalebox{0.8}{45.09} & \scalebox{0.8}{49.91} \\
        
        & \scalebox{0.8}{MAPE} & \scalebox{0.8}{\secondres{15.52\%}} &-    & \scalebox{0.8}{\boldres{15.50\%}} & \scalebox{0.8}{24.38\%} & \scalebox{0.8}{24.56\%} & \scalebox{0.8}{20.65\%} & \scalebox{0.8}{25.62\%} & \scalebox{0.8}{22.51\%} & \scalebox{0.8}{25.63\%} & \scalebox{0.8}{25.82\%} & \scalebox{0.8}{23.23\%} & \scalebox{0.8}{20.18\%} & \scalebox{0.8}{24.26\%} \\
        \hline
        % \multirow{2}{*}{$1^{st}$ Count} 
        \multicolumn{2}{c}{\scalebox{0.8}{$1^{\textit{st}}$ Count}} & \scalebox{0.8}{\boldres{9}} & \scalebox{0.8}{0} & \scalebox{0.8}{\secondres{4}} & \scalebox{0.8}{0} & \scalebox{0.8}{0} & \scalebox{0.8}{0} & \scalebox{0.8}{0} & \scalebox{0.8}{0} & \scalebox{0.8}{0} & \scalebox{0.8}{0} & \scalebox{0.8}{2} & \scalebox{0.8}{0} & \scalebox{0.8}{0}  \\
        \toprule
    \end{tabular}
    \end{small}
  \end{threeparttable}
}
\end{table*}

\subsection{Experimental Setup}
\subsubsection{\textbf{Datasets}} 
In contrast to the region-specific, small-scale spatiotemporal datasets commonly used in prior studies, we conducted training and evaluation on large-scale, public, real-world datasets. The datasets used in our experiments are summarized in Table S1 of the Supplementary Materials. They span five traffic-related metrics: traffic flow, taxi demand, bicycle trajectories, traffic speed, and traffic index and across major metropolitan areas (such as New York City, Chicago, Los Angeles, Shanghai, and Shenzhen). The pre-training dataset encompasses 10,110 regions and 352,796 time points, resulting in a total of 151,089,924 spatio-temporal observations~\cite{li2024opencity}. These datasets vary in the number of variables, sampling frequency, spatial scale, temporal duration, and overall data volume.
During the testing phase, we evaluated the model on datasets excluded from training, allowing for an assessment of its generalization capability across diverse traffic prediction scenarios.
Further details regarding the experimental datasets are provided in the supplementary material.
\subsubsection{\textbf{Baseline}}
We have carefully selected the latest and advanced models in spatio-temporal prediction community as our baselines, which including: (1) Recent Spatio-Temporal foundation models: OpenCity~\cite{li2024opencity}, UniST~\cite{yuan2024unist}; (2) Mamba-based models: SpoT-Mamba~\cite{choi2024spot}, STG-Mamba~\cite{li2024stg}; (3) Attention-Based models: ASTGCN~\cite{guo2019attention}, STWA~\cite{cirstea2022towards}, PDFormer~\cite{jiang2023pdformer}; and (4) GNNs-based models: STGCN~\cite{yu2018spatio}, TGCN~\cite{zhao2019t}, GWN~\cite{wu2019graph}, MTGNN~\cite{wu2020connecting}, STSGCN~\cite{song2020spatial}. Further Baseline details are provided in the Supplementary Material. 

\subsubsection{\textbf{Evaluation Metrics and Implementation Details}}
We employ widely used regression metrics to evaluate model performance, including Mean Absolute Error (MAE), Root Mean Square Error (RMSE), and Mean Absolute Percentage Error (MAPE), where lower values indicate better performance. To ensure fair and comprehensive comparisons, we follow the experimental setup of~\cite{li2024opencity}. All training and testing are conducted on a server equipped with 6× NVIDIA GeForce RTX A6000 48GB GPUs. Our largest model contains 12M parameters.
For the long-term traffic forecasting task, we set both the historical and future time spans to 288 time steps. The patch length and stride are set to 12. The regularization weights are fixed across all datasets as $\alpha = 1$ and $\beta = 0.5$, while the fusion weights are set to $w_1 = 0.4$ and $w_2 = 0.6$ during training.
For all baselines, we adopt the hyperparameter configurations as reported in their original papers or officially released implementations. Further implementation details are provided in the Supplementary Materials. Note that UniST is designed specifically for grid-based data and, therefore, does not produce results on graph-based datasets.

\subsection{In-distribution Prediction (RQ1)}
\vskip -0.05in
\subsubsection{\textbf{Evaluation Setups}}
We evaluate the adaptability of our proposed Damba-ST in an in-distribution scenario, where a large portion of the data is used for training, and the test set is held out for evaluation. Following the setup in~\cite{li2024opencity}, we assess the performance of the compared models in a long-term prediction setting.
This is a challenging task due to the evolving temporal distribution shifts over longer time horizons. We directly use the pre-trained models of Damba-ST and OpenCity for testing, while other domain-specific baselines are trained end-to-end on their respective datasets. 
\subsubsection{\textbf{Results}}
As shown in Table~\ref{tab:In-distribution Prediction}, the results indicate that our proposed Damba-ST consistently outperforms the baselines across most evaluation metrics. Notably, we observed that almost all specialized models underperformed in the long-term traffic flow prediction task (i.e., on datasets CAD12-2, CAD8-2, and CAD8-1). This underperformance can primarily be attributed to these models' tendency to overfit historical spatio-temporal patterns, limiting their ability to generalize to future temporal shifts. In contrast, our proposed Damba-ST and OpenCity, both foundation models trained on large-scale datasets, have learned universal spatio-temporal knowledge from diverse data sources and exhibit robust generalization capabilities in this challenging task. 
\subsection{Zero-shot Prediction (RQ2)}
\vskip -0.05in
\begin{table*}[!htbp]
    \caption{Results of Zero-shot Prediction. Performance comparison between Damba-ST under the zero-shot setting and baseline models under the full-shot setting across three increasingly challenging cross-domain tasks: \textit{Cross-Region, Cross-City, and Cross-Task}.}
    \vskip -0.1in
    \label{tab:zero-shot}
    \renewcommand{\arraystretch}{0.85} 
    \centering
    \resizebox{\textwidth}{!}{
    \begin{threeparttable}
    \begin{small}
    \renewcommand{\multirowsetup}{\centering}
    \setlength{\tabcolsep}{1pt}
    \begin{tabular}{c c c c c c c c c c c c c c c c}
        \toprule
        \multicolumn{2}{c}{\multirow{3}{*}{\scalebox{0.8}{Dataset}}} & \multirow{3}{*}{\scalebox{0.8}{Metrics}} & 
        \multicolumn{3}{c}{\scalebox{0.8}{Foundation Models}}&\multicolumn{2}{c}{\scalebox{0.8}{Mamba-based Models}} & \multicolumn{3}{c}{\scalebox{0.8}{Attention-based Models}} & \multicolumn{5}{c}{\scalebox{0.8}{GNNs-based Models}} \\
       \cmidrule(lr){4-6}\cmidrule(lr){7-8}\cmidrule(lr){9-11}\cmidrule(lr){12-16}
       
        & & & \scalebox{0.85}{$\textbf{Damba-ST}$} & \scalebox{0.8}{UniST} & \scalebox{0.8}{OpenCity} & \scalebox{0.8}{STGMamba} & \scalebox{0.8}{SpoTMamba}  & \scalebox{0.8}{PDFormer} & \scalebox{0.8}{STWA} & \scalebox{0.8}{ASTGCN} & \scalebox{0.8}{STSGCN} & \scalebox{0.8}{MTGNN} & \scalebox{0.8}{GWN} & \scalebox{0.8}{TGCN} & \scalebox{0.8}{STGCN} \\
        
        & & & \scalebox{0.8}{$\textbf{(Ours)}$} & \scalebox{0.8}{~\cite{yuan2024unist}} & \scalebox{0.8}{~\cite{li2024opencity}} & \scalebox{0.8}{~\cite{li2024stg}} & \scalebox{0.8}{~\cite{choi2024spot}} &  \scalebox{0.8}{~\cite{jiang2023pdformer}} & \scalebox{0.8}{~\cite{cirstea2022towards}} & \scalebox{0.8}{~\cite{guo2019attention}} & \scalebox{0.8}{~\cite{song2020spatial}} & \scalebox{0.8}{~\cite{wu2020connecting}} & \scalebox{0.8}{~\cite{wu2019graph}} & \scalebox{0.8}{~\cite{zhao2019t}} & \scalebox{0.8}{~\cite{yu2018spatio}} \\
        \toprule
        
        \multirow{9}{*}{\rotatebox{90}{Region}} & \multirow{3}{*}{\scalebox{0.8}{CAD3}} 
        &  \scalebox{0.8}{MAE} &\scalebox{0.8}{\boldres{14.21}} & - & \scalebox{0.8}{\secondres{15.88}} &\scalebox{0.8}{21.35}& \scalebox{0.8}{22.73}  & \scalebox{0.8}{20.28} & \scalebox{0.8}{21.65} & \scalebox{0.8}{23.60} & \scalebox{0.8}{21.88} & \scalebox{0.8}{17.59} & \scalebox{0.8}{16.94} & \scalebox{0.8}{19.56} & \scalebox{0.8}{20.24}\\
        
        & & \scalebox{0.8}{RMSE} & \scalebox{0.8}{\boldres{24.75}} & - & \scalebox{0.8}{\secondres{27.03}} & \scalebox{0.8}{36.92} & \scalebox{0.8}{37.12} & \scalebox{0.8}{36.43} & \scalebox{0.8}{37.55} & \scalebox{0.8}{39.35} & \scalebox{0.8}{34.52} & \scalebox{0.8}{28.92} & \scalebox{0.8}{28.81} & \scalebox{0.8}{30.82} & \scalebox{0.8}{34.34}\\
        
        & &\scalebox{0.8}{MAPE} & \scalebox{0.8}{\boldres{20.03\%}} & - & \scalebox{0.8}{\secondres{21.94\%}} & \scalebox{0.8}{25.97\%} & \scalebox{0.8}{26.93\%} & \scalebox{0.8}{25.19\%} & \scalebox{0.8}{26.85\%} & \scalebox{0.8}{41.22\%} & \scalebox{0.8}{30.20\%} & \scalebox{0.8}{25.22\%} & \scalebox{0.8}{22.98\%} & \scalebox{0.8}{28.15\%} & \scalebox{0.8}{25.33\%}\\
        \cline{2-16}
        
        & \multirow{3}{*}{\scalebox{0.8}{CAD5}}
        &  \scalebox{0.8}{MAE} & \scalebox{0.8}{\secondres{10.77}} &-    & \scalebox{0.8}{11.09} & \scalebox{0.8}{12.94} & \scalebox{0.8}{13.27}  & \scalebox{0.8}{12.89} & \scalebox{0.8}{14.43} & \scalebox{0.8}{12.58} & \scalebox{0.8}{13.87} & \scalebox{0.8}{11.70} & \scalebox{0.8}{\boldres{10.69}} & \scalebox{0.8}{13.07} & \scalebox{0.8}{13.76} \\
        
        & & \scalebox{0.8}{RMSE} & \scalebox{0.8}{\boldres{18.27}} &-    & \scalebox{0.8}{\secondres{18.96}} & \scalebox{0.8}{23.52} & \scalebox{0.8}{23.12} & \scalebox{0.8}{21.18} & \scalebox{0.8}{24.14} & \scalebox{0.8}{21.23} & \scalebox{0.8}{22.32} & \scalebox{0.8}{20.30} & \scalebox{0.8}{19.75} & \scalebox{0.8}{21.56} & \scalebox{0.8}{27.49} \\
        
        & & \scalebox{0.8}{MAPE} & \scalebox{0.8}{\secondres{26.31\%}} &-    & \scalebox{0.8}{27.90\%} & \scalebox{0.8}{30.06\%} & \scalebox{0.8}{29.77\%}  & \scalebox{0.8}{28.82\%} & \scalebox{0.8}{29.34\%} & \scalebox{0.8}{30.56\%} & \scalebox{0.8}{32.84\%} & \scalebox{0.8}{26.75\%} & \scalebox{0.8}{\boldres{25.98\%}} & \scalebox{0.8}{32.65\%} & \scalebox{0.8}{32.89\%} \\
        \cline{2-16}
        
        & \multirow{3}{*}{\scalebox{0.8}{PEMS07M}} 
        &  \scalebox{0.8}{MAE} & \scalebox{0.8}{4.53} &-    & \scalebox{0.8}{4.50} & \scalebox{0.8}{4.65} & \scalebox{0.8}{4.77} & \scalebox{0.8}{4.62} & \scalebox{0.8}{4.54} & \scalebox{0.8}{\secondres{4.39}} & \scalebox{0.8}{4.56} & \scalebox{0.8}{4.52} & \scalebox{0.8}{\boldres{4.17}} & \scalebox{0.8}{4.88} & \scalebox{0.8}{4.44} \\
        & & \scalebox{0.8}{RMSE} & \scalebox{0.8}{8.33} &-    & \scalebox{0.8}{8.21} & \scalebox{0.8}{8.92} & \scalebox{0.8}{8.43} & \scalebox{0.8}{8.36} & \scalebox{0.8}{8.57} & \scalebox{0.8}{8.21} & \scalebox{0.8}{\secondres{8.05}} & \scalebox{0.8}{8.06} & \scalebox{0.8}{\boldres{7.84}} & \scalebox{0.8}{8.38} & \scalebox{0.8}{8.37} \\
        & & \scalebox{0.8}{MAPE} & \scalebox{0.8}{12.57\%} &-    & \scalebox{0.8}{\secondres{12.20\%}} & \scalebox{0.8}{13.28\%} & \scalebox{0.8}{14.02\%}  & \scalebox{0.8}{13.74\%} & \scalebox{0.8}{12.91\%} & \scalebox{0.8}{12.58\%} & \scalebox{0.8}{12.70\%} & \scalebox{0.8}{13.11\%} & \scalebox{0.8}{\boldres{11.46\%}} & \scalebox{0.8}{13.94\%} & \scalebox{0.8}{12.59\%} \\
        \hline\\[-1.5mm]\hline
        
        \multirow{6}{*}{\rotatebox{90}{City}} & \multirow{3}{*}{\scalebox{0.8}{TrafficSH}} 
        & \scalebox{0.8}{MAE} & \scalebox{0.8}{\boldres{0.51}} & \scalebox{0.8}{1.26} & \scalebox{0.8}{\secondres{0.55}} & \scalebox{0.8}{1.24} & \scalebox{0.8}{1.13} & \scalebox{0.8}{0.77} & \scalebox{0.8}{1.42} & \scalebox{0.8}{0.69} & \scalebox{0.8}{1.66} & \scalebox{0.8}{0.81} & \scalebox{0.8}{0.76} & \scalebox{0.8}{1.79} & \scalebox{0.8}{1.60} \\
        
        & & \scalebox{0.8}{RMSE} & \scalebox{0.8}{\boldres{0.85}} & \scalebox{0.8}{1.72} & \scalebox{0.8}{\boldres{0.85}} & \scalebox{0.8}{2.83} & \scalebox{0.8}{2.97} & \scalebox{0.8}{1.23} & \scalebox{0.8}{2.49} & \scalebox{0.8}{1.09} & \scalebox{0.8}{3.33} & \scalebox{0.8}{1.26} & \scalebox{0.8}{1.39} & \scalebox{0.8}{2.65} & \scalebox{0.8}{3.14} \\
        & & \scalebox{0.8}{MAPE} & \scalebox{0.8}{\boldres{7.83\%}} & \scalebox{0.8}{8.69\%} & \scalebox{0.8}{\secondres{8.01\%}} & \scalebox{0.8}{10.03\%} & \scalebox{0.8}{10.39\%} & \scalebox{0.8}{8.34\%} & \scalebox{0.8}{9.92\%} & \scalebox{0.8}{8.05\%} & \scalebox{0.8}{9.33\%} & \scalebox{0.8}{8.29\%} & \scalebox{0.8}{9.23\%} & \scalebox{0.8}{17.75\%} & \scalebox{0.8}{8.04\%} \\
        \cline{2-16}

        & \multirow{3}{*}{\scalebox{0.8}{CHI-TAXI}}
        & \scalebox{0.8}{MAE} & \scalebox{0.8}{\secondres{2.05}} & \scalebox{0.8}{4.26} & \scalebox{0.8}{\boldres{1.91}} & \scalebox{0.8}{3.88} & \scalebox{0.8}{4.25} & \scalebox{0.8}{4.03} & \scalebox{0.8}{3.70} & \scalebox{0.8}{3.28} & \scalebox{0.8}{4.87} & \scalebox{0.8}{3.27} & \scalebox{0.8}{3.56} & \scalebox{0.8}{4.02} & \scalebox{0.8}{3.09} \\
        & & \scalebox{0.8}{RMSE} & \scalebox{0.8}{\secondres{4.53}} & \scalebox{0.8}{13.57} & \scalebox{0.8}{\boldres{4.42}} & \scalebox{0.8}{11.32} & \scalebox{0.8}{13.94} & \scalebox{0.8}{12.82} & \scalebox{0.8}{11.49} & \scalebox{0.8}{10.32} & \scalebox{0.8}{14.40} & \scalebox{0.8}{9.87} & \scalebox{0.8}{11.27} & \scalebox{0.8}{11.70} & \scalebox{0.8}{9.54} \\
        & & \scalebox{0.8}{MAPE} & \scalebox{0.8}{42.31\%} & \scalebox{0.8}{45.29\%} & \scalebox{0.8}{\secondres{40.07\%}} & \scalebox{0.8}{45.73\%} & \scalebox{0.8}{46.72\%} & \scalebox{0.8}{44.42\%} & \scalebox{0.8}{42.52\%} & \scalebox{0.8}{42.82\%} & \scalebox{0.8}{104.64\%} & \scalebox{0.8}{\boldres{39.38\%}} & \scalebox{0.8}{41.31\%} & \scalebox{0.8}{60.25\%} & \scalebox{0.8}{42.47\%} \\
        \hline\\[-1.5mm]\hline
        
        \multirow{3}{*}{\rotatebox{90}{Task}}& \multirow{3}{*}{\scalebox{0.8}{NYC-BIKE}} 
        & \scalebox{0.8}{MAE} & \scalebox{0.8}{\boldres{6.09}} & \scalebox{0.8}{7.36} & \scalebox{0.8}{\secondres{6.32}} & \scalebox{0.8}{6.94} & \scalebox{0.8}{7.23} & \scalebox{0.8}{7.33} & \scalebox{0.8}{7.97} & \scalebox{0.8}{6.44} & \scalebox{0.8}{6.85} & \scalebox{0.8}{6.48} & \scalebox{0.8}{7.61} & \scalebox{0.8}{7.75} & \scalebox{0.8}{8.01} \\
        & & \scalebox{0.8}{RMSE} & \scalebox{0.8}{\boldres{11.46}} & \scalebox{0.8}{12.58} & \scalebox{0.8}{11.60} & \scalebox{0.8}{12.95} & \scalebox{0.8}{13.46} & \scalebox{0.8}{13.01} & \scalebox{0.8}{14.35} & \scalebox{0.8}{11.62} & \scalebox{0.8}{11.98} & \scalebox{0.8}{\secondres{11.49}} & \scalebox{0.8}{13.56} & \scalebox{0.8}{13.49} & \scalebox{0.8}{13.94} \\
        & & \scalebox{0.8}{MAPE} & \scalebox{0.8}{\secondres{60.27\%}} & \scalebox{0.8}{65.39\%} & \scalebox{0.8}{70.06\%} & \scalebox{0.8}{64.76\%} & \scalebox{0.8}{63.29\%} & \scalebox{0.8}{65.44\%} & \scalebox{0.8}{64.19\%} & \scalebox{0.8}{63.90\%} & \scalebox{0.8}{68.44\%} & \scalebox{0.8}{61.52\%} & \scalebox{0.8}{\boldres{58.41\%}} & \scalebox{0.8}{85.27\%} & \scalebox{0.8}{63.02\%} \\
        \hline
        % \multirow{2}{*}{$1^{st}$ Count} 
        \multicolumn{3}{c}{\scalebox{0.8}{$1^{\textit{st}}$ Count}} & \scalebox{0.8}{\boldres{9}} & \scalebox{0.8}{0} & \scalebox{0.8}{3} & \scalebox{0.8}{0} & \scalebox{0.8}{0} & \scalebox{0.8}{0} & \scalebox{0.8}{0} & \scalebox{0.8}{0} & \scalebox{0.8}{0} & \scalebox{0.8}{1} & \scalebox{0.8}{\secondres{6}} & \scalebox{0.8}{0} & \scalebox{0.8}{0} \\
        \toprule
        
    \end{tabular}
    \end{small}
  \end{threeparttable}
}
\end{table*}
\subsubsection{\textbf{Evaluation Setups}} 
Foundation models (Damba-ST, UniST, OpenCity) are evaluated in a strict zero-shot manner: they are applied directly to each target test set with \emph{no} fine-tuning, and none of these targets are used during pre-training. In contrast, all other baselines are trained on the target data (full-shot).
We evaluate Damba-ST’s zero-shot generalization at three progressively harder levels: (i) \textit{Cross-Region}: generalization to unseen regions within the same city; (ii) \textit{Cross-City}: transfer to entirely new cities by leveraging knowledge learned elsewhere; and (iii) \textit{Cross-Task}: forecasting previously unseen traffic metrics.

\subsubsection{\textbf{Results}} 

As shown in Table~\ref{tab:zero-shot}, Damba-ST ranks top-2 on most datasets and metrics without any target-specific retraining, indicating that it captures transferable spatio-temporal regularities across regions, cities, and tasks. The zero-retraining property shortens deployment time and reduces compute, making Damba-ST a practical cold-start solution. Across the \textit{Cross-Region} and \textit{Cross-City} settings, it consistently achieves best or second-best performance, demonstrating that a single pre-trained model can generalize well to unseen regions and cities with diverse traffic conditions.
For the \textit{Cross-Task} setting, although bicycle trajectories are not included during pretraining, Damba-ST still maintains excellent MAE and RMSE, confirming its versatility and ability to adapt to previously unseen traffic metrics in real-world scenarios.
Specialized baselines (e.g., GWN, MTGNN) surpass Damba-ST are reasonable under these setups, since those models are \emph{trained on the target domain} with full access to target labels and are tuned to the characteristics of a single dataset. In other words, they act as strong in-domain experts. By contrast, Damba-ST is evaluated in a strict zero-shot manner with frozen parameters; Our goal is not to beat every target-specific expert, but to deliver consistently high performance across all regions, cities, and tasks with one unified, general-purpose model.

\begin{figure}[!h]
\centerline{\includegraphics[width=0.49\textwidth]{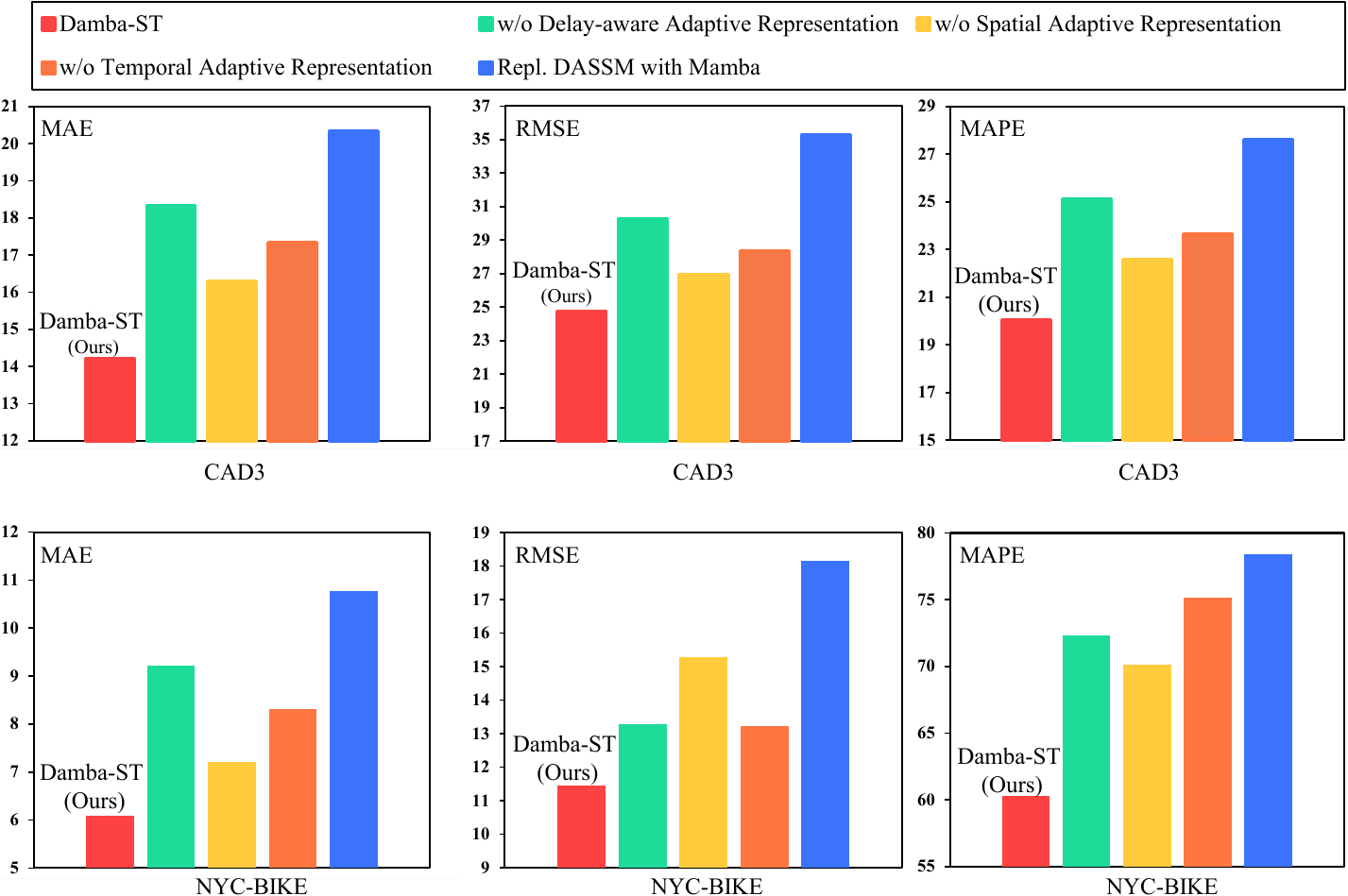}}
% \vskip -0.1in
\caption{Ablation Study on the CAD3 and NYC-BIKE Datasets}.
\label{fig:Ablations}
\vskip -0.3in
\end{figure}
\subsection{Ablation Study (RQ3)}
% \vskip -0.05in
To validate the effectiveness of the key components in Damba-ST, we conducted an ablation study under zero-shot settings using the CAD3 and NYC-BIKE datasets.
The ablation study includes the following variants: replacing \emph{(Repl.)} the DASSM module with vanilla Mamba in three DASSM variants, and individually removing the $\text{DASSM}_\mathrm{S}$, $\text{DASSM}_\mathrm{T}$, and $\text{DASSM}_\mathrm{D}$.
As shown in Fig.~\ref{fig:Ablations}, we can make the following observations:
\begin{itemize}[leftmargin=*]
\item \textbf{\emph{Repl.} DASSM: } When replacing DASSM with Mamba, the model reverts to fusion training and fails to capture the heterogeneity of the data. This results in the most significant performance degradation, as the model loses its ability to learn domain-adaptive representations.
\item \textbf{\emph{w/o} $\text{DASSM}_\mathrm{S}$ (Spatial View):} Removing the Spatial Adaptive Representations leads to a noticeable increase in error metrics. By considering the structural information of the traffic graph, the model can capture generalized spatial-topology-related traffic patterns, which are crucial for cross-region generalization.
\item \textbf{\emph{w/o} $\text{DASSM}_\mathrm{T}$ (Temporal View):} Removing the Temporal Adaptive Representation results in a significant degradation of performance, as the temporal view provides essential information about the series's own dynamics. This enables the model to capture time-dependent traffic patterns, which are essential for making accurate predictions.
\item \textbf{\emph{w/o} $\text{DASSM}_\mathrm{D}$ (ST-delay View):} Removing the Delay-Aware Adaptive Representations causes a significant performance drop. The ST-delay View provides knowledge of spatio-temporal delay dependencies, which are crucial for real-world traffic prediction, as it captures the delayed interactions between spatial and temporal factors.
\end{itemize}

\subsection{Efficiency Analysis and Deployment Practicality (RQ4) }
\vskip -0.05in
Damba-ST theoretically offers significant advantages over Transformer-based spatio-temporal foundation models in terms of training complexity, inference complexity, and memory usage, as shown in Table~\ref{tab:Deployment}.
Specifically, the computational complexity\footnote{For fairness, we only compare the attention module in Transformer-based models with the SSM module in Damba-ST. A more detailed complexity analysis is provided in the Supplementary Material.} of Damba-ST scales linearly with the input sequence length $L$, whereas Transformer-based models typically exhibit quadratic complexity with respect to $L$. This makes Damba-ST more efficient and memory-friendly.
We empirically validate these advantages on real-world datasets from \textit{Training Efficiency} and \textit{Deployment Practicality}.
\begin{table}[!h]
\caption{Complexity Analysis and Deployment Practicality. $L$ is input sequence length, $D$ is the hidden dimension, $L \gg D$.}
\vskip -0.1in
\label{tab:Deployment}
\centering
\begin{threeparttable}
\begin{small}
\renewcommand{\multirowsetup}{\centering}
  \setlength{\tabcolsep}{2pt}
\begin{tabular}{cccc}
\toprule
\multirow{2}{*}{Comparison}  & \multicolumn{2}{c}{Transformer-based} & Mamba-based \\
\cmidrule(lr){2-3}\cmidrule(lr){4-4}
& UniST~\cite{yuan2024unist} & Opencity~\cite{li2024opencity} & \textbf{Damba-ST}(Ours)\\ 

\toprule
Training Complexity  & $\mathcal{O}(L^2D)$  & $\mathcal{O}(L^2D)$  & \backg$\mathcal{O}(3LD^2)$  \\
Inference Complexity & $\mathcal{O}(LD)$  & $\mathcal{O}(LD)$  & \backg$\mathcal{O}(3D^2)$  \\
Memory               & $\mathcal{O}(L^2)$ & $\mathcal{O}(L^2)$   & \backg$\mathcal{O}(3LD)$    \\  
\hline\\[-1.5mm]\hline
MAE & 2.94 & 1.74   & \backg1.51    \\  
RMSE & 7.88 & 3.80 & \backg3.22    \\  
Cost (seconds) & 1.3s  & 1.5s   & \backg0.9s    \\  
\bottomrule
\end{tabular}
\end{small}
  \end{threeparttable}
\end{table}
\subsubsection{\textbf{Training Efficiency}} 
\begin{figure}[!h]
\centerline{\includegraphics[width=0.43\textwidth]{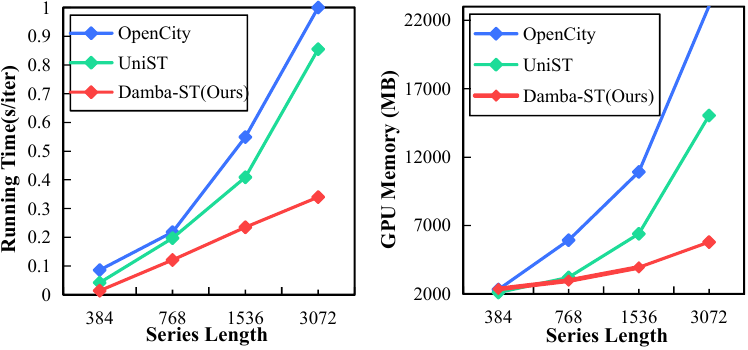}}
% \vskip -0.1in
\caption{Efficiency analysis of running time and GPU memory in long-term prediction. Damba-ST scales linearly with the series length.}
\label{fig:effciency}
\vskip -0.25in
\end{figure}

As shown in Fig.~\ref{fig:effciency}, we compare the proposed Damba-ST against existing spatio-temporal foundation models Unist~\cite{yuan2024unist} and OpenCity~\cite{li2024opencity} in terms of Running Time (i.e., Training Speed) and GPU memory usage for long-term forecasting tasks, where the lookback window length increases quadratically from $\{384, 768, 1536, 3072\}$. The figure illustrates that transformer-based spatiotemporal foundation models, such as UniST and OpenCity, exhibit quadratic complexity, resulting in high memory usage and slower training speeds. In contrast, the Mamba-based Damba-ST offers a more efficient trade-off between performance, training speed, and memory usage. Damba-ST consistently shows superior efficiency in both training time and GPU memory usage across varying series lengths, achieving a favorable balance between performance and computational cost.
% \vspace{-12pt}
\subsubsection{\textbf{Deployment Practicality}}
Given that all experimental datasets are sourced from real-world urban environments, we further evaluate the practical efficiency of Damba-ST under deployment settings. To simulate a realistic application scenario, we test Damba-ST on the CHI-TAXI dataset, which was not included in the pre-training, by performing next-day traffic prediction using data from the previous day. The model is deployed on an NVIDIA GeForce RTX A6000 48GB GPU.

As shown in Table~\ref{tab:Deployment}, Damba-ST not only achieves the best performance in terms of MSE and RMSE but also demonstrates remarkable inference efficiency, completing each prediction in just 0.9 seconds. This rapid response time underscores Damba-ST’s suitability for real-time urban applications and highlights its potential as a robust and deployable spatio-temporal foundation model, offering a win-win solution in both prediction accuracy and computational efficiency.

\begin{figure*}[!htbp]
\begin{center}
\subfigure[$w_1, w_2$ on CAD3 dataset (cross-region)]{
\includegraphics[width=0.3\linewidth]{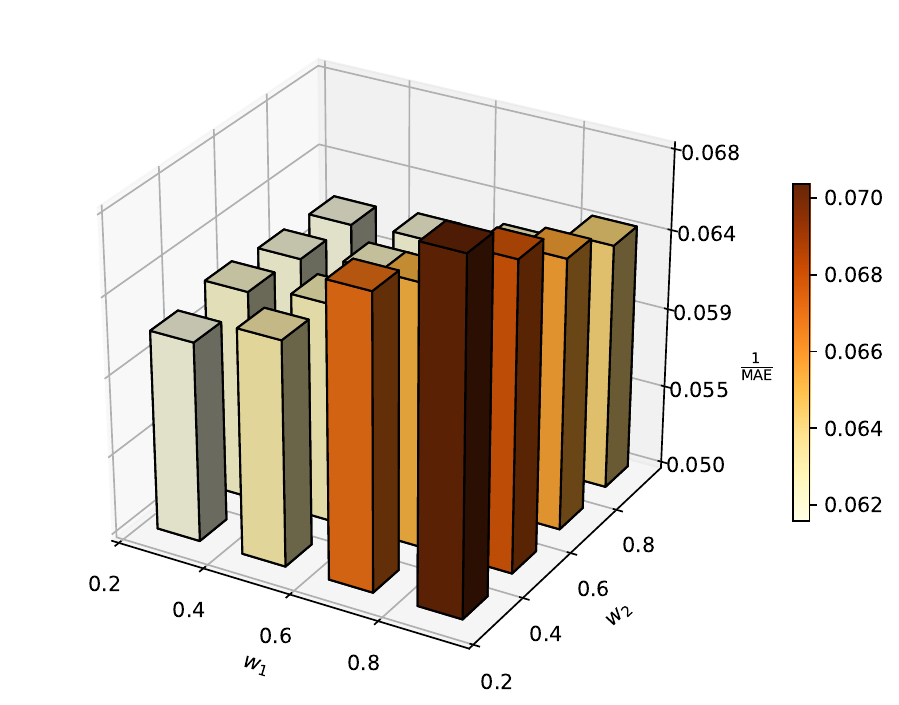}
\label{fig:}
}
\subfigure[$w_1, w_2$ on TrafficSH dataset (cross-city)]{
\includegraphics[width=0.3\linewidth]{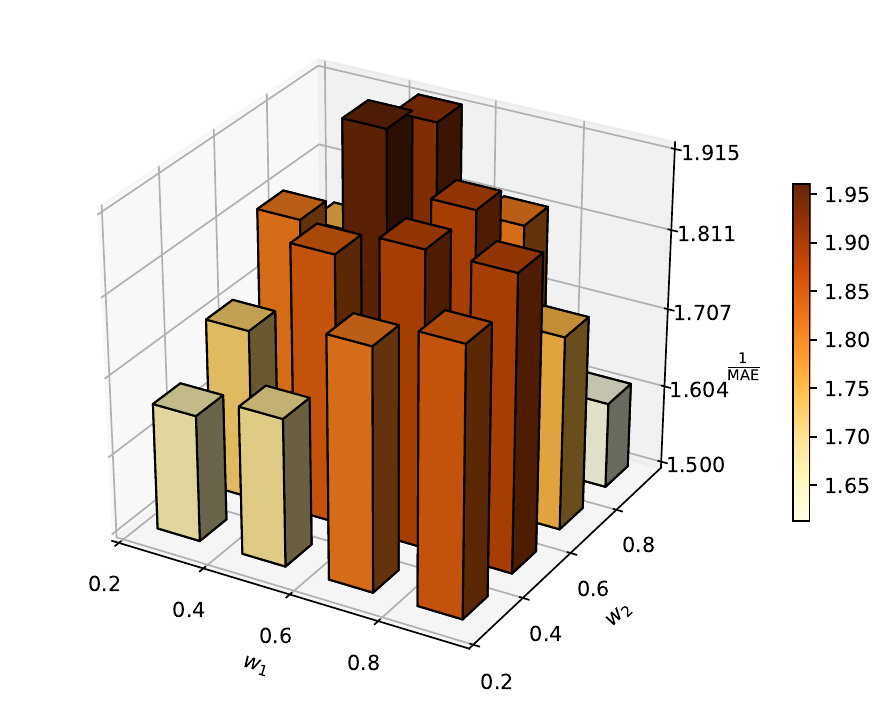}
\label{fig:}
}
\subfigure[$w_1, w_2$ on NYC-BIKE dataset (cross-task)]{
\includegraphics[width=0.3\linewidth]{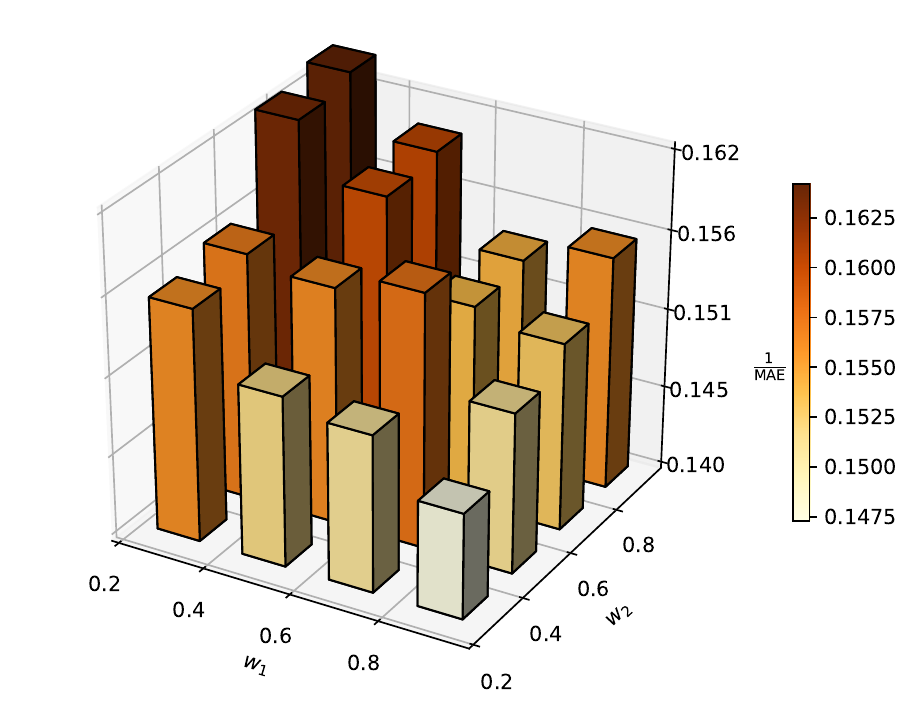}
\label{fig:}
}
\vskip -0.1in
\subfigure[$\alpha, \beta$ on CAD3 dataset (cross-region)]{
\includegraphics[width=0.3\linewidth]{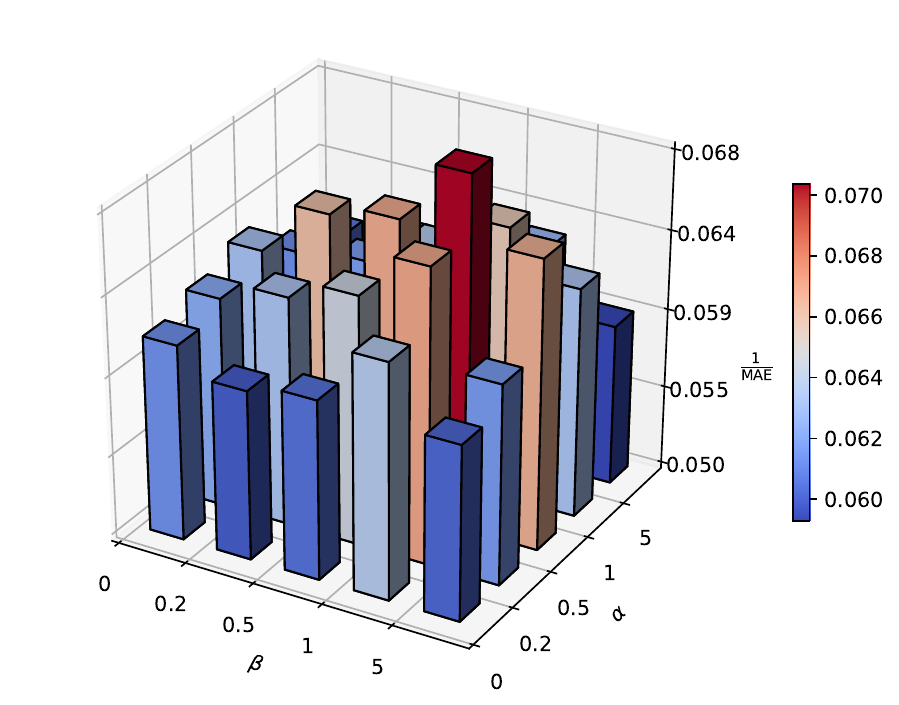}
\label{fig:}
}
\subfigure[$\alpha, \beta$ on TrafficSH dataset (cross-city)]{
\includegraphics[width=0.3\linewidth]{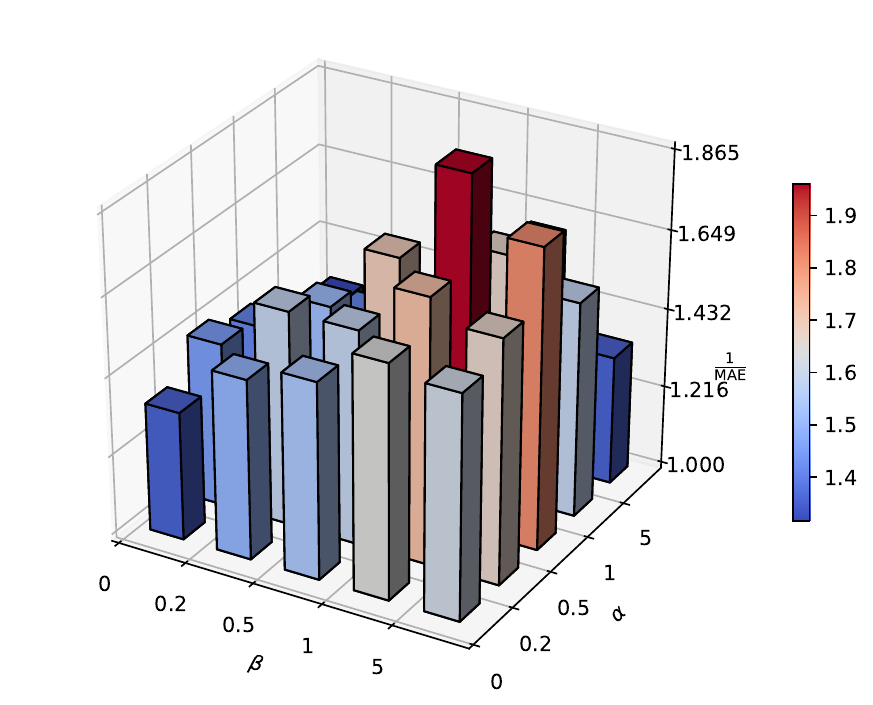}
\label{fig:}
}
\subfigure[$\alpha, \beta$ on NYC-BIKE dataset (cross-task)]{
\includegraphics[width=0.3\linewidth]{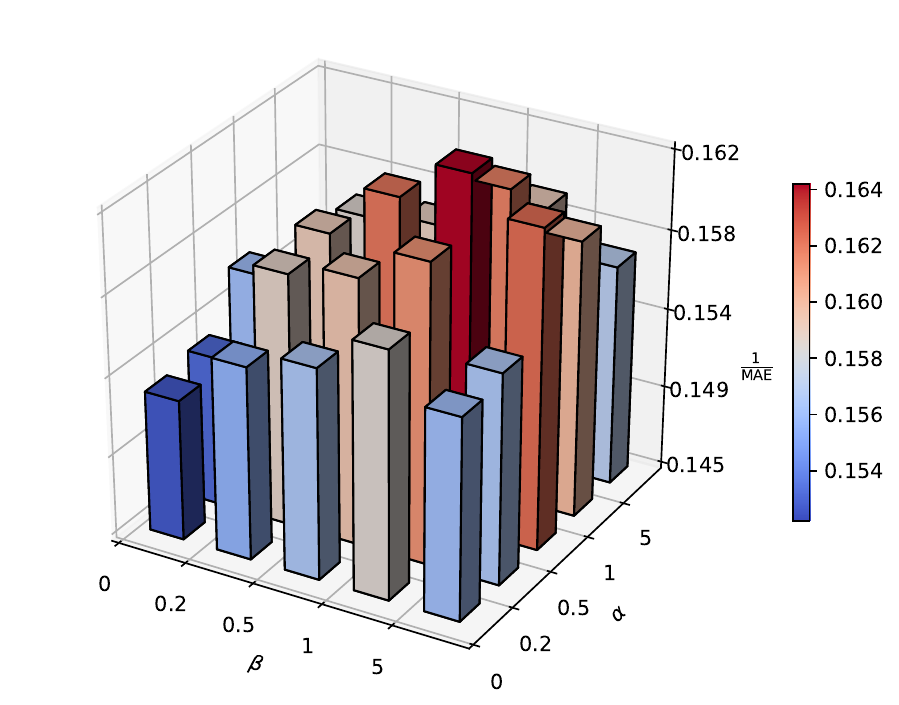}
\label{fig:}
}
 \vskip -0.1in
\caption{Hyperparameter Analysis. }
\vskip -0.3in
\label{fig:Hyperparameter Analysis}
\end{center}
\end{figure*}

% \subsection{Scalability(RQ5)}
% \vspace{-6pt}
\subsection{Hyperparameter Sensitivity Analysis (RQ5)}
\vskip -0.05in
We evaluate the hyperparameter sensitivity of Damba-ST on three zero-shot datasets. Specifically, we examine two sets of hyperparameters: (1) the fusion weights $w_1$ and $w_2$, which control the trade-off between the Discriminative Representation and Common Representation in the DASSM fusion layer (as shown in Eq~\eqref{eq:fusion}); and (2) the regularization weights $\alpha$ and $\beta$, which govern the similarity constraint term in the objective function (as shown in Eq~\eqref{eq:finalloss}).
The evaluation metric is MAE.

As shown in Fig.~\ref{fig:Hyperparameter Analysis}(a)-(c), we perform grid searches over fusion weights $w_1, w_2 \in \{0.2, 0.4, 0.6, 0.8\}$. We find that the fusion weights $ w_1$ and $ w_2$ should be carefully selected based on the differences between the training and testing data. Specifically, when testing on an unseen region, assigning a higher weight to the Discriminative Representation weights $w_1$ typically yields better results (as shown in Fig.~\ref{fig:Hyperparameter Analysis}(a)). Conversely, when testing on an unseen task—which is more challenging than a cross-region task—the model’s overall generalization capability becomes crucial, and thus, a higher weight on the Common Representation weights $w_2=0.8$ tends to produce superior performance (as shown in Fig.~\ref{fig:Hyperparameter Analysis}(b)).

As shown in Fig.~\ref{fig:Hyperparameter Analysis} (d)-(f),We perform grid searches of $\alpha, \beta \in \{0, 0.2, 0.5, 1, 5\}$. We find that increasing $\alpha, \beta$ generally improves performance, with the best results at $\alpha=1, \beta=0.5$. Beyond this point, the performance stabilizes or slightly decreases, indicating that while regularization is crucial, excessive regularization can have a slight negative impact on performance.

\section{Related Work}
\label{sec:literature}
\vskip -0.05in
\subsection{\textbf{Spatio-temporal Foundation Models.}} 
\vskip -0.05in
Inspired by the significant advancements in foundation models for NLP and CV, a notable line of work has explored spatiotemporal foundation models. Some LLM-based models, such as CityGPT~\cite{feng2024citygpt}, CityBench~\cite{feng2024citybench}, and UrbanGPT~\cite{li2024urbangpt}, have demonstrated proficiency in addressing language-based urban tasks. LLMs have been employed to generate descriptions of urban-related images, thus aiding downstream tasks such as urban scene understanding~\cite{feng2024citygpt}. While the integration of LLMs into urban systems shows great promise, it is important to recognize that spatio-temporal data is not inherently generated by language; the modality gap limits the direct application of LLMs to spatio-temporal tasks. Recent efforts have focused on training spatio-temporal foundation models from scratch. For instance, UniST~\cite{yuan2024unist} was developed for grid-based urban scenarios, allowing it to generalize to new situations without additional training. Similarly, OpenCity~\cite{li2024opencity} integrates the Transformer architecture with graph neural networks to model complex spatio-temporal dependencies, benefiting from pretraining on large-scale data. Despite these advancements, the high computational requirements of the Transformer architecture pose challenges for computational efficiency and real-world deployment. Moreover, the fusion training methods induce challenges in generalization due to spatio-temporal heterogeneity. Consequently, there is an urgent need to develop more efficient and generalized models that can effectively capture universal spatio-temporal patterns across diverse, heterogeneous spatio-temporal scenarios.

\subsection{\textbf{Mamba-based Spatio-Temporal Model.}} 
\vskip -0.05in
Inspired by the superior performance of Mamba models across various domains~\cite{qu2024survey}, including language modeling~\cite{lieber2024jamba}, speech analysis~\cite{quan2024multichannel}, and recommender systems~\cite{qu2024ssd4rec}, researchers have recently started exploring the applicability of Mamba for spatio-temporal forecasting and spatial-temporal graph learning~\cite{choi2024spot, li2024stg, shao2024st, yuan2024st}. For instance, STG-Mamba~\cite{li2024stg} treats the spatio-temporal graph evolution process as a system and employs a selective mechanism to focus on relevant latent features. SpoT-Mamba~\cite{choi2024spot} performs Mamba scanning across both spatial and temporal dimensions to capture corresponding dependencies. ST-Mamba integrates CNNs with Mamba for traffic flow estimation. While these models leverage Mamba’s computational efficiency and strong performance for specific tasks and domains, to date, no research has explored Mamba as a spatio-temporal foundation model capable of generalizing across diverse urban scenarios.

\section{Conclusion}
\label{sec:conclusion}
\vskip -0.05in
Training urban spatio-temporal models that generalize across cities is challenging due to data heterogeneity and computational complexity. To tackle these issues, we introduce Damba-ST, which addresses spatio-temporal heterogeneity through domain adapters, three scan strategies, and a domain-adaptive selective state space model to capture both shared and domain-specific features. Our experiments demonstrate that Damba-ST achieves state-of-the-art performance, with superior zero-shot capabilities, enabling seamless deployment in new environments without retraining.

\section{Acknowledgement}
\label{sec:ack}
\vskip -0.05in

The research described in this paper has been partially supported by the National Natural Science Foundation of China (project no. 62433016 and 62537001), General Research Funds from the Hong Kong Research Grants Council (project no. PolyU 15207322, 15200023, 15206024, and 15224524), internal research funds from Hong Kong Polytechnic University (project no. P0042693, P0048625, and P0051361). This work was supported by computational resources provided by The Centre for Large AI Models (CLAIM) of The Hong Kong Polytechnic University.

\section{AI-Generated Content Acknowledgement}
During the preparation of this work, the authors used ChatGPT to polish the language. After using this tool, the authors reviewed and edited the content as needed and take full responsibility for the final version of the manuscript.
\balance
\bibliographystyle{IEEEtran}
\bibliography{references/references}

\clearpage

\end{document}